\documentclass[10pt,twocolumn,letterpaper]{article}

\usepackage{cvpr}

\usepackage{macros}
\usepackage{placeins} %
\usepackage{fontawesome5}

\newcommand{\method}{BFD}
\newcommand{\bfit}[1]{\textbf{\textit{(#1)}}}

\makeatletter
\renewcommand{\maketitle}{\twocolumn[
  \begin{center}
    \vspace*{0.5cm}
    \hrule height 2pt
    \vspace{0.6cm}
    {\LARGE \bfseries \@title \par}
    \vspace{0.6cm}
    \hrule height 1pt
    \vspace{0.6cm}
    {\large \@author \par}
    
    \vspace{0.8cm}
  \end{center}
]}
\makeatother

\title{Bi-Orthogonal Factor Decomposition for Vision Transformers}

\author{
\textbf{Fenil R. Doshi}\textcolor{secondary}{$^\star$$^{a,b}$}\quad
\textbf{Thomas Fel}\textcolor{secondary}{$^\star$$^b$}\quad
\textbf{Talia Konkle}\textcolor{secondary}{$^{a,b}$}\quad
\textbf{George A. Alvarez}\textcolor{secondary}{$^{a,b}$} \\[0.5em]
{\normalsize \textcolor{secondary}{$^a$}Dept. of Psychology, Harvard University \qquad \textcolor{secondary}{$^b$}Kempner Institute, Harvard University}
\\
{
\small 
\href{https://www.fenildoshi.com/bfd-transformers/}{\raisebox{-2.0pt}{\includegraphics[height=10pt]{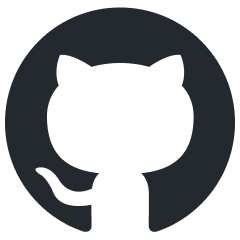}}
\texttt{fenildoshi.com/bfd-transformers}}
}
}

\begin{document}
\pagenumbering{arabic}
\pagestyle{plain}

\twocolumn[{%
\renewcommand\twocolumn[1][]{#1}%
\maketitle
\vspace{-3.0em}
\begin{center}
    \centering
    \captionsetup{type=figure}
    \includegraphics[width=0.9\linewidth]{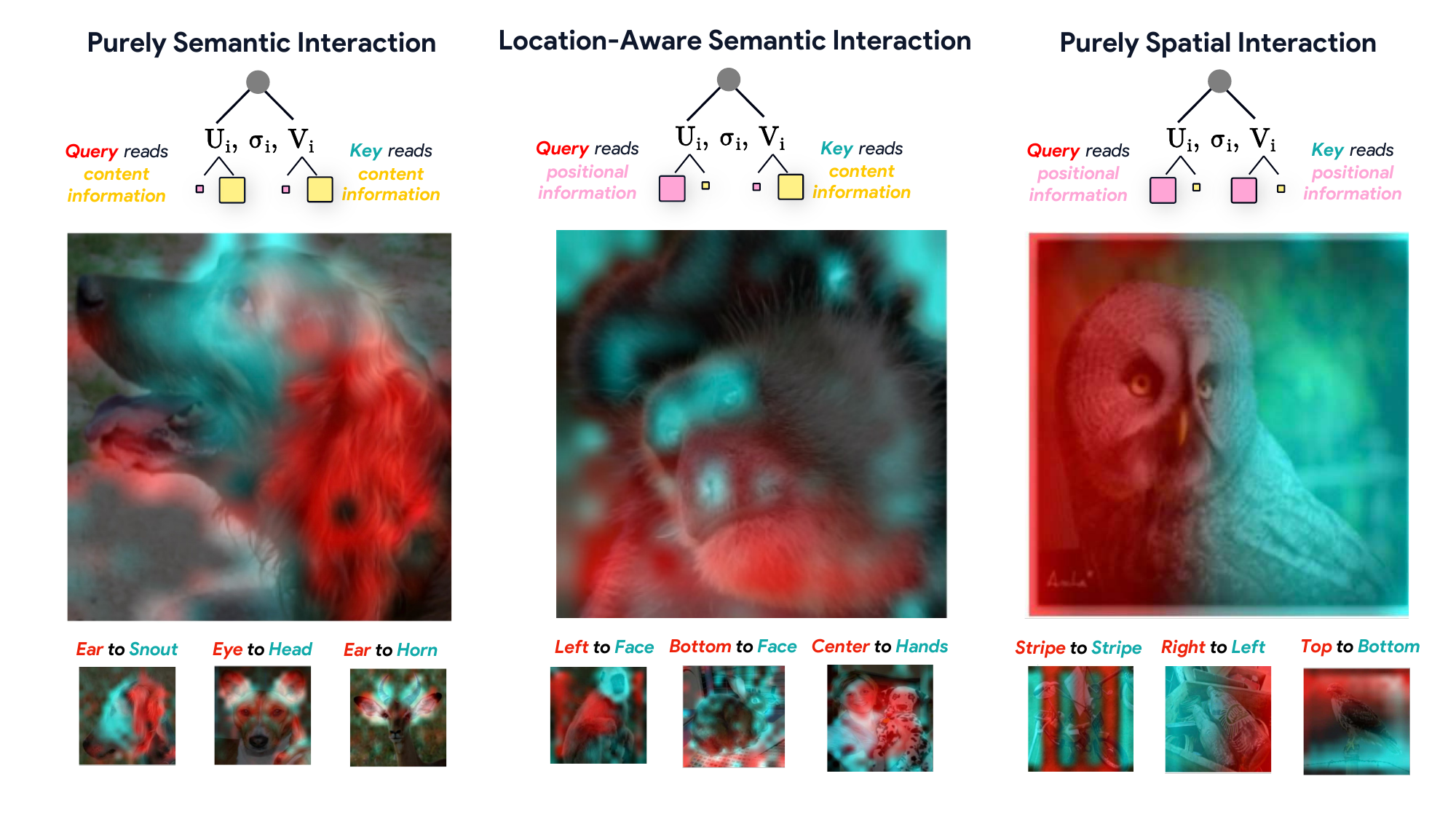}
    \captionof{figure}{\textbf{Content and Position Interactions in DINOv2.} Representative bi-orthogonal modes in DINOv2 whose activation patterns illustrate three different interaction. Left: Content–content modes activate on semantic interactions such as object parts, revealing part-to-part or part-to-whole correspondences. Middle: Content–position modes show localization-aware semantic interactions, where semantic features are modulated by spatial context. Right: Position–position modes activate on purely spatial interaction, like left–right flows, top–bottom variations, and Fourier-like patterns, highlighting geometric organization without semantic selectivity. These examples show how different interaction types contribute to information flow in attention.
    }
    \label{fig:teaser}
\end{center}%
}]

\renewcommand{\thefootnote}{}
\footnotetext{\textcolor{secondary}{$^\star$} Denotes equal contribution.}
\renewcommand{\thefootnote}{\arabic{footnote}} 

\vspace{-3mm}
\begin{abstract}
\vspace{0mm}
Self-attention is the central computational primitive of Vision Transformers, yet we lack a principled understanding of what information attention mechanisms exchange between tokens. 
Attention maps describe where weight mass concentrates; they do not reveal whether queries and keys trade position, content, or both.
We introduce Bi-orthogonal Factor Decomposition (\method), a two-stage analytical framework: first, an ANOVA-based decomposition statistically disentangles token activations into orthogonal positional and content factors; second, SVD of the query-key interaction matrix $~\bm{QK}^\tr$ exposes bi-orthogonal modes that reveal how these factors mediate communication.
After validating proper isolation of position and content, we apply \method~to state-of-the-art vision models and uncover three phenomena.
\bfit{i}~Attention operates primarily through content. Content-content interactions dominate attention energy, followed by content-position coupling. DINOv2 allocates more energy to content-position than supervised models and distributes computation across a richer mode spectrum.
\bfit{ii}~Attention mechanisms exhibit specialization: heads differentiate into content-content, content-position, and position-position operators, while singular modes within heads show analogous specialization. 
\bfit{iii}~DINOv2's superior holistic shape processing emerges from intermediate layers that simultaneously preserve positional structure while contextually enriching semantic content. 

Overall, \method~exposes how tokens interact through attention and which informational factors -- positional or semantic -- mediate their communication, yielding practical insights into vision transformer mechanisms.

\end{abstract}

\section{Introduction}
\label{sec:intro}
Vision Transformers (ViTs)~\cite{vaswani2017attention,dosovitskiy2020image,oquab2023dinov2,darcet2023vision} have become the workhorses of modern computer vision, powering recognition systems deployed in medical and other safety-critical settings~\cite{wang2022artificial,alecu2022can} and serving as the backbone of large-scale generative models~\cite{zheng2025diffusion,zhang2023tale} used by millions ~\cite{chui2023economic}.
Understanding these systems is therefore both a practical imperative and a scientific opportunity~\cite{lecun2015deep,serre2006learning}. From a scientific perspective, ViTs are among the most scalable~\cite{alabdulmohsin2023getting,zhai2022scaling,dehghani2023scaling} and capable vision models available~\cite{zhai2023sigmoid,he2022masked,touvron2022deit,caron2021emerging}; 
it is natural to ask whether their internal computations share structure with human visual processing~\cite{fel2022aligning,sucholutsky2023getting,muttenthaler2022human,kornblith2019similarity, kriegeskorte2015deep, chung2021neural} and, if so, where such similarities emerge~\cite{cadieu2014deep,linsley2023performance,conwell2022can, konkle2022self}? 
To answer those question, we turn to the interpretability field~\cite{doshivelez2017rigorous,jacovi2020towards,gilpin2018explaining}, which has proposed a range of theories~\cite{elhage2022superposition,park2023linear,costa2025flat} and methods~\cite{zeiler2013visualizing,kim2018interpretability,bricken2023monosemanticity} for understanding model internals. 
Early interpretability efforts centered on attribution~\cite{zeiler2013visualizing,fong2017meaningful,smilkov2017smoothgrad,selvaraju2017gradcam,fel2021sobol,muzellec2023gradient,petsiuk2018rise}. These methods primarily aim to reveal where a model attends or which regions of an input most influence its prediction. While such tools offer valuable intuition, they face well-documented limitations~\cite{adebayo2018sanity,kim2021hive,fel2021cannot,sixt2020explanations,sixt2022users,nguyen2021effectiveness}. They often highlight broad image regions without clarifying what visual patterns drive the decision, nor how these signals are combined internally. 
In response, concept-based methods emerged~\cite{kim2018interpretability,ghorbani2017interpretation,fel2023craft,fel2023holistic,zhang2021invertible,vielhaben2023multi,shukor2024implicit,parekh2024concept,dreyer2025mechanistic,kowal2024understanding}, shifting focus from spatial attribution to semantic characterization. Rather than showing \emph{where} the model looks, these approaches aim to identify \emph{what} it has learned -- isolating human-interpretable ``visual atoms''~\cite{ullman2016atoms,serre2006learning} or latent concepts within its representations~\cite{bricken2023monosemanticity}. Concept activation vectors~\cite{kim2018interpretability}, network dissection~\cite{bau2017network} or feature visualization~\cite{olah2017feature,fel2023unlocking,hamblin2024feature} exemplify this paradigm. Such analyses have revealed, for instance, that deep networks spontaneously form units selective for texture, object parts, or higher-order scene semantics~\cite{fel2024understanding,kowal2024visual}.
Yet, despite their insights, these techniques remain representation-centric: they explain \emph{what} information is encoded, not \emph{how} that information arises. They stop short of elucidating the circuits~\cite{lepori2023neurosurgeon} or computations~\cite{schmidhuber2002speed} that produce these representations -- a question that has become central to modern interpretability.

To uncover these internal computations, attention mechanisms present a natural entry point. Self-attention is a core component of ViTs as it mediates communication between tokens and thus provides an explicit interface for information exchange. 
Existing analyses have mostly interpreted attention through its spatial patterns -- visualizing which tokens attend to which others across layers~\cite{xu2015show,choi2016retain}. However, these maps reflect only the surface structure of interactions: they indicate the amount of attention exchanged, not the content of that exchange or the reasons for it~\cite{wiegreffe2019attention,bastings2020elephant,akula2022attention}. This gap has fueled growing skepticism over whether raw attention maps genuinely constitute explanations~\cite{bibal2022attention,tutek2020staying}.
Recently, however, a more mechanistic turn has emerged. Rather than treating attention weights as explanatory endpoints, researchers probe the underlying query-key interactions directly via spectral structure~\cite{pan2024dissecting}. These methods aim to identify dominant interaction modes and relate them to concrete behaviors, echoing circuit-level analyses pioneered in large language models~\cite{valeriani2023geometry,song2023uncovering,bushnaq2025stochastic,chrisman2025identifying} (e.g., induction heads~\cite{olsson2022context}, copying mechanisms~\cite{mcdougall2023copy}, compositional reasoning~\cite{bai2025can,lepori2024break}).
Yet, even when dominant modes are identified, their nature often remains ambiguous: are these modes driven primarily by positional geometry, by semantic content, or by global trends such as the mean activation direction? Without a principled factor attribution framework~\cite{song2023uncovering} (one that disentangles these potential sources) it remains unclear whether the apparent contextual integration observed in deeper layers reflects genuinely content-driven computation or merely the imprint of positional structure.
This motivates our first question.
\begin{question} What informational factors -- positional, semantic, or global -- mediate token communication through attention, and how does this composition vary across layers and between supervised vs. self-supervised models?
\label{question:1}
\end{question}
Understanding what information flows between tokens naturally leads to asking \emph{how} this flow is organized at the network level. The architecture of multi-head attention introduces a second source of ambiguity here as each head could, in principle, implement a distinct computational role (tracking spatial relations, aggregating semantics, ...) supporting a modular view of attention. Alternatively, information may be distributed across heads, with individual heads contributing only partial, entangled signals to a collective computation. 
These contrasting modular versus distributed hypotheses carry deep implications for interpretability: the former supports circuit-level mechanistic explanations at the head level, while the latter implies that per-head analyses may be fundamentally incomplete.
This leads to our second question.
\begin{question}
Do attention heads and their constituent singular modes exhibit functional specialization into distinct informational operators (content-content, content-position, position-position), and is this specialization consistent across layers and models?
\label{question:2}
\end{question}
Understanding the organization of attention naturally raises a further question: what determines it?
If head specializations exist, are they intrinsic to the architecture or induced by the learning paradigm? Evidence suggests that training objectives and data statistics strongly influence representational geometry and inductive biases~\cite{caron2021emerging,bardes2021vicreg,zbontar2021barlow}.
Self-supervised models such as DINOv2~\cite{oquab2023dinov2,darcet2023vision} display greater holistic shape processing abilities~\cite{geirhos2018imagenet,doshi2025visual} than their supervised counterparts, implying that training may alter how information is integrated within attention. Identifying the nature and location of these changes could clarify why self-supervision yields more invariant and globally coherent representations.
This motivates our third question.
\begin{question}
What structural properties of intermediate-layer representations distinguish self-supervised from supervised models, and how do these properties—specifically positional preservation and content enrichment—enable holistic shape processing?
\label{question:3}
\end{question}
\paragraph{Contributions.} 
To address these questions, we build upon the analytical paradigm of~\cite{song2023uncovering,merullo2024talking} (originally developed for LLMs) and contextualize it for vision. Specifically, 
\begin{itemize}
    \item \textbf{We introduce \textit{Bi-Orthogonal Factor Decomposition} (\method).} A theoretical framework that couples a statistical factorization of activations with a spectral decomposition of the attention interaction matrix. 
    \item \textbf{Quantifying information flow in attention.}  \method~quantifies how much attention energy flows through positional versus content-based interactions. Content-based modes (content-content and content-position) carry the majority of energy in both architectures, confirming that contextual integration reflects semantic computation rather than positional structure. DINOv2 dedicates more energy to content-position coupling and operates through higher-rank mode spectra, suggesting richer interactions.
    \item \textbf{Functional specialization of attention heads.}  
    Across models and layers, heads differentiate into content–content, content–position, and position–position operators. Within individual heads, singular modes show analogous specialization, establishing a quantitative notion of head function.
    \item \textbf{Dual preservation as the signature of holistic shape processing.} DINOv2 simultaneously preserves 2D spatial topology and enriches semantic content through intermediate layers. 
    We find that the activation similarities that usually exhibit block-diagonal structure disappear when examining content alone.
    This reveals that content progressively enriches by integrating both positional and semantic information.
    Supervised models, by contrast, collapse spatial structure to quasi-1D by mid-depth, precluding this localized hierarchical integration.

\end{itemize}

To address these questions, we build upon the analytical paradigm pioneered by \citet{song2023uncovering} for language models and contextualize it to ViT. 

\begin{figure*}[t]
  \vspace{-9mm}
  \centering
  \includegraphics[width=\linewidth]{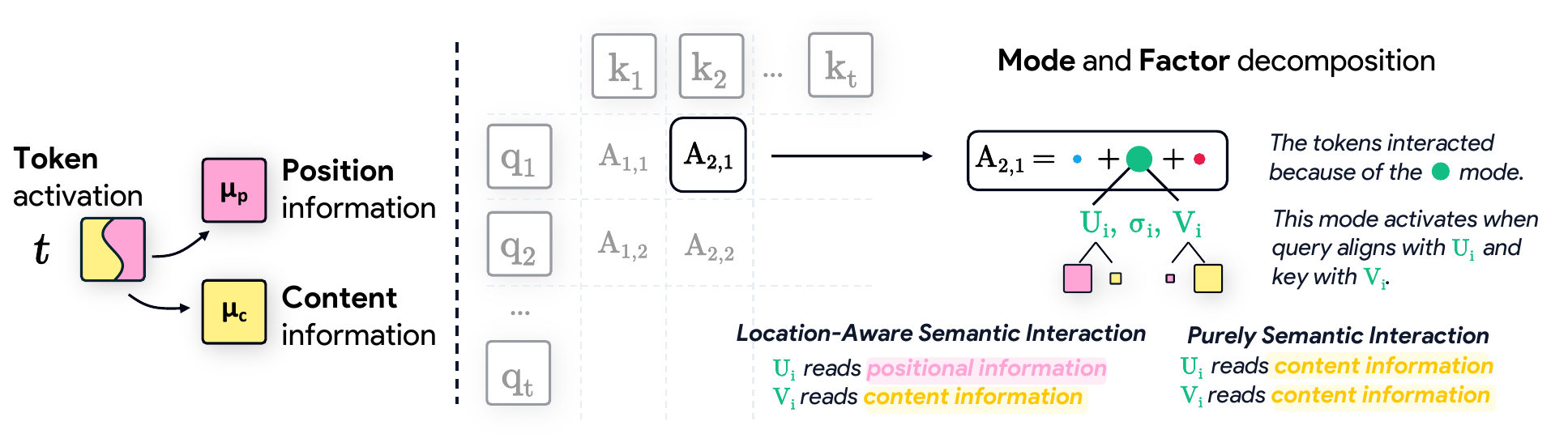}
  \vspace{-5mm}
  \caption{\textbf{Bi-orthogonal Factor Decomposition (BFD).}
      Each token activation is decomposed into positional ($\mu_p$) and content ($\mu_c$) components.
      The bilinear interaction between these components is then analyzed through a singular value decomposition, 
      yielding biorthogonal mode pairs $(U_i, V_i)$ that explain why two tokens interact. 
      A mode is activated when the query aligns with $U_i$ and the key with $V_i$. For example, when $U_i$ aligns with a token’s positional component while $V_i$ aligns with another token’s content component, the mode expresses a location-aware semantic interaction, effectively mixing where something is with what it is. This decomposition exposes which paired directions enable tokens to interact and clarifies whether their interaction arises from positional or content factors or both.
    }
  \vspace{-3mm}
  \label{fig:method}
\end{figure*}

\section{Theoretical Framework}
\label{sec:method}

We present Bi-orthogonal Factor Decomposition (\method), which couples statistical factorization of activations with spectral decomposition of attention (\cref{fig:method}). The framework proceeds in two stages: first, we adapt ANOVA-based factorization to disentangle spatial topology from semantic content in vision representations; second, we perform SVD on the query-key interaction matrix to reveal how these factors drive token communication through bi-orthogonal modes.

\paragraph{Preliminaries.}Let $\f$ a ViT that partition an image $\x \in \SX$ into $P$ non-overlapping patches, each linearly projected to form a token embedding. The network admits a series of $L$ intermediate representations 
$$
\A_{\ell} = \f_{\ell}(\x), ~~\ell \in \{1, \ldots, L\} \quad \text{where} \quad \A_{\ell} \in \mathbb{R}^{P \times d},
$$
stacks the $d$-dimensional embeddings for all $P$ tokens at layer $\ell$. We denote the embedding of token $p$ at layer $\ell$ as $\A_{\ell}^{(p)} \in \mathbb{R}^d$. 
The architecture alternates between element-wise transformations (feedforward blocks) and token-mixing operations performed by multi-head self-attention. Each block's output is added to the residual stream, forming the standard Transformer architecture~\cite{vaswani2017attention,dosovitskiy2020image}.
Crucially, positional information must be injected into tokens; without it, attention would be permutation-invariant~\cite{xu2024permutation} and thus insensitive to spatial structure~\cite{cordonnier2019relationship}. Most ViTs achieve this by adding a learned positional embedding $\bm{\gamma}_p \in \mathbb{R}^d$ to each token at the input layer: $\A_{0}^{(p)} = \x^{(p)} + \bm{\gamma}_p,$
where $\x^{(p)}$ denotes the linearly projected patch embedding. This additive positional bias introduces spatial grounding and enables attention to mix tokens in a position-dependent manner.
The central questions of this work concern how tokens interact through attention, and what information—positional or semantic—is exchanged during this interaction. To address this, we decompose the Transformer computation into two interpretable stages: \bfit{i} a statistical factorization of activations that disentangles position and content, and \bfit{ii} a bi-orthogonal spectral decomposition of attention that exposes its communication modes.

\paragraph{Statistical Factorization.}
Positional embeddings enter the model additively and linearly at the input layer, suggesting that subsequent activations retain a linearly recoverable positional component. We exploit this property to construct a statistically orthogonal decomposition of activations into three factors: global mean average (layer effect), positional bias (token position effect), and content residual (token idiosyncratic information). This two-ways ANOVA style decomposition provides a principled separation of signal sources~\cite{seber2009multivariate}. 

\begin{definition}[Vision Transformer Spatial-Content Factorization]
\label{def:scd}
Let $\mathbb{E}_{\x}$ denote expectation over images and $\mathbb{E}_{\x,p}$ expectation over images and token indices. For any vision transformer $\f$ at layer $\ell$, every patch embedding admits the unique additive decomposition:
\vspace{-2mm}
\begin{equation*}
\f_{\ell}^{(p)}(\x) = \layer + \position + \content, ~~\text{where}
\end{equation*}
\vspace{-4mm}
\begin{equation*}
\begin{cases}
\layer = \mathbb{E}_{(\x,p)}(\f_{\ell}^{(p)}(\x)) & \text{(Layer effect)} \\
\position = \mathbb{E}_{\x}(\f_{\ell}^{(p)}(\x)) - \layer & \text{(Positional effect)} \\
\content = \f_{\ell}^{(p)}(\x) - \layer - \position & \text{(Content residual)}
\end{cases}
\end{equation*}
\end{definition}

Where the content residual $\content$ isolates input-specific semantic information and is statistically orthogonal to both global and positional components by construction. Specifically, in expectation over the joint distribution of $(\x,p)$ we have:
\begin{align*}
\mathbb{E}_{\x,p}\left(\layer^{\top}\position\right) &= 
\mathbb{E}_{\x,p}\left(\layer^{\top}\content\right) \\
& =\mathbb{E}_{\x}\left({\position}{\ }^{\top}\content\right) = 0.
\end{align*}
Essentially, this factorization uses dataset and spatial marginals to isolate factors that are already implicated by the architecture. Extensions to finer-grained marginalizations, for instance, conditioning on image level activations or semantic classes are conceptually straightforward and constitute promising directions for future work.

Having decomposed activations into interpretable factors, we now require a complementary decomposition to analyze how these factors interact through attention. Specifically, we seek to decompose the query-key interaction in a manner that reveals which paired directions drive token communication at each layer and head.

\paragraph{Bi-orthogonal Decomposition}

We now analyze the mechanism that mediates information exchange: self-attention. For a given head, the attention is then defined as
$$
\bm{Y} = \operatorname{softmax}\left(\frac{(\A\bm{W}_Q)(\A\bm{W}_K)^\tr}{\sqrt{d}}\right)\V,
$$
where we recall that \(\A \in \mathbb{R}^{P \times d}\) stacks token activations for \(P\) positions. 
Specifically, the interaction between queries and keys occurs through the bilinear form \((\A\bm{W}_Q)(\A\bm{W}_K)^\tr\), which encapsulates all pairwise relations before normalization. Following the spectral analyses of~\cite{merullo2024talking,song2023uncovering,pan2024dissecting}, we characterize this operator via singular value decomposition:
\begin{align*}
(\A \bm{W}_{Q})(\A \bm{W}_{K})^\tr = 
\A\bm{W}_{Q}\bm{W}_{K}^\tr\A^\tr = \A \bm{W} \A^\tr
\end{align*}
With $\bm{W} \equiv \bm{W}_{Q}\bm{W}_{K}^\tr$ the interaction matrix. We decompose this matrix using SVD $\bm{W} = \bm{U}\bm{\Sigma}\bm{V}$.
Each triplet \((\bm{u}_i, \sigma_i, \bm{v}_i)\) defines a bi-orthogonal mode~\cite{dieudonne1953biorthogonal,kaplan1956biorthogonality}: left and right singular vectors \(\bm{u}_i\) and \(\bm{v}_i\) are orthonormal bases of the query and key subspaces, and \(\sigma_i\) quantifies their coupling strength. Crucially, the bi-orthogonality structure ensures that mode \(i\) operates independently: queries aligned with \(\bm{u}_i\) communicate exclusively with keys aligned with \(\bm{v}_i\), while \(\bm{u}_i\) cannot interact with \(\bm{v}_j\) for \(j \neq i\). This orthogonal decomposition of the attention interaction thus partitions information flow into independent (non-interfering) communication channels.
\begin{definition}[Bi-orthogonal Mode Decomposition.]
\label{def:bmd}
Given a token representation $\A$ and the interaction matrix decomposition $\bm{W} \equiv \bm{W}_{Q}\bm{W}_{K}^\tr = \bm{U}\bm{\Sigma}\bm{V}$, we define the projected codes
\[
\z^Q = \bm{A}\bm{U} , \qquad 
\z^K = \bm{A}\bm{V}.
\]
The pair \((\bm{z}^Q_i, \bm{z}^K_i)\) represents the token activations aligned with the \(i\)-th communication mode, whose strength is modulated by \(\sigma_i\).
\end{definition}
This decomposition is complete: the query-key interaction matrix can be exactly reconstructed by summing over all modes,
\vspace{-4mm}
\begin{equation*}
\vspace{-2mm}
\bm{QK}^\top = \A\bm{W}\A^\top = \sum_{i=1}^{d} \bm{z}^Q_i \sigma_i (\bm{z}^K_i)^\top,
\end{equation*}
ensuring that the spectral decomposition partitions attention energy without information loss.

\noindent With both the factor decomposition of activations and the mode decomposition of attention now established, we can unify these frameworks to attribute each communication mode to its underlying informational source (position, content, or global bias), thereby exposing what information flows through each attention channel.
\noindent With both the factor decomposition of activations and the mode decomposition of attention now established, we can unify these frameworks to attribute each communication mode to its underlying informational source (position, content, or global bias), thereby exposing what information flows through each attention channel.
\paragraph{\method: Coupling Factors and Modes.} Combining the activation factorization with the bi-orthogonal decomposition enables us to attribute each communication mode to its underlying informational factors: position, content, or global bias. 
Recall that each layer's activations decompose using the three factors $\F = (\layer, \position, \content)$:
\vspace{-2mm}
\begin{equation*}
\A_{\ell} = \layer + \position + \content = \sum_{\factor \in \F} \factor.
\end{equation*}
Critically, the coupled decomposition is also complete: the query-key interaction can be exactly reconstructed by summing over all factors and modes,
\vspace{-2mm}
\begin{equation*}
\bm{QK}^\top = \sum_{\bm{\mu} \in \F} \sum_{i=1}^{d} (\bm{\mu} \bm{u}_i) \sigma_i (\bm{\mu} \bm{v}_i)^\top,
\end{equation*}
We can now project each factor of layer $\ell$ onto the query and key modes and measure its associated energy on the $i$-th mode. Formally,

\vspace{-2mm}
\begin{equation*}
\energy^{(i)}_{\cdot} = \|\bm{z}^Q_{\cdot,i} \cdot \sigma_i \cdot \bm{z}^K_{\cdot,i} \|_2^2,
\end{equation*}
where $\bm{z}^Q_{\cdot,i}$ is the projection of a factor onto query mode $\bm{u}_i$ (e.g., $\position \bm{U}_i$ for position), similarly for keys. The normalized energy:

\vspace{-5mm}
\begin{equation*}
    \bar{\energy}^{(i)}_{\cdot}
    = \mathbb{E}_{\x}
      \left(
        \frac{
          \energy^{(i)}_{\cdot}(\x)
        }{
          \sum_{j} \energy^{(j)}_{\cdot}(\x)
        }
      \right),
    \qquad
    \sum_{i} \bar{\energy}^{(i)}_{\cdot} = 1.
\end{equation*}

quantifies the relative dominance of each factor within mode $i$ (e.g, $\bar{\energy}_{\texttt{c}}^{(i)}$ is the relative energy of the content information for the $i$-th mode).
Aggregating across modes and heads yields layer-level summaries that characterize the informational profile of attention at different depths.
These quantities operationalize statements such as "this head is content-dominated" or "this layer pivots from position to content," providing a quantitative backbone linking representational factorization to emergent specialization in Vision Transformers.

\section{Results}
\label{sec:results}

\paragraph{Experimental Setup.}
We apply our framework to two representative Vision Transformers: a supervised ViT-B/16 from\texttt{timm} library~\cite{wightman2019pytorch} and a self-supervised DINOv2-B/14 with registers~\cite{darcet2023vision}. Both models share identical architectural parameters: 12 transformer blocks, $16\times16$ patch tokenization in the supervised ViT-B/16  and $14\times14$ patch tokenization yielding 196 and 256 spatial tokens, plus a class token (and 4 additional register tokens for DINOv2). We compute the statistical factorization (Definition~\ref{def:scd}) and bi-orthogonal mode decomposition (Definition~\ref{def:bmd})across all 12 layers for both architectures, analyzing 5,000 images sampled from the ImageNet validation set. 

\begin{figure}[t]
    \vspace{-6mm}
    \centering
    \includegraphics[width=0.99\linewidth]{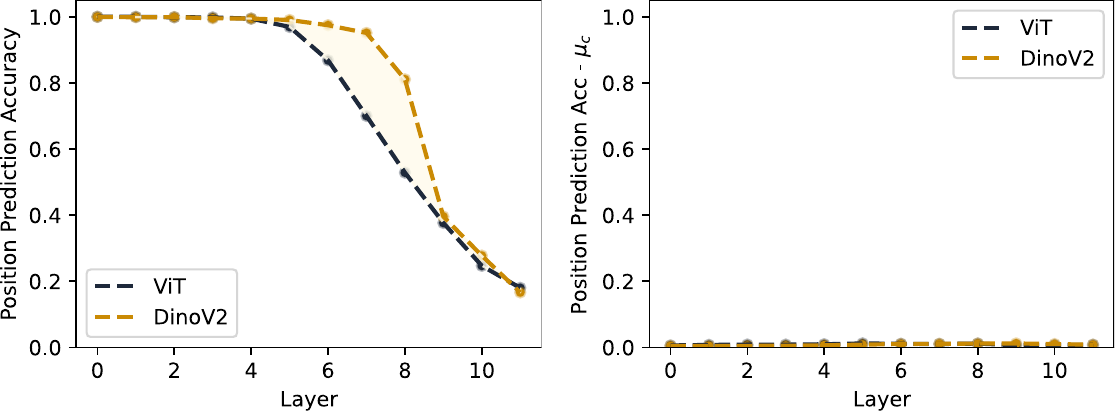}
    \vspace{-3mm}
    \caption{\textbf{Validating Factor Isolation.} Linear probes decode spatial coordinates from factorized representations across layers. \textit{Left:} Position is successfully decoded from the raw block activations in both architectures, with DINOv2 maintaining higher accessibility through intermediate layers before both converge to 20\% accuracy at depth.
     \textit{Right:} Content factors $\content$ contain negligible positional information (chance-level decoding), confirming that the factorization removes positional information and achieves statistical orthogonality. The asymmetric preservation trajectory reveals that self-supervised training maintains linearly accessible positional structure deeper into the network than supervised learning.}
     \vspace{-3mm}
    \label{fig:sanity_check_position}
\end{figure}

\begin{figure*}[t]
  \vspace{-9mm}
  \centering
  \includegraphics[width=\linewidth]{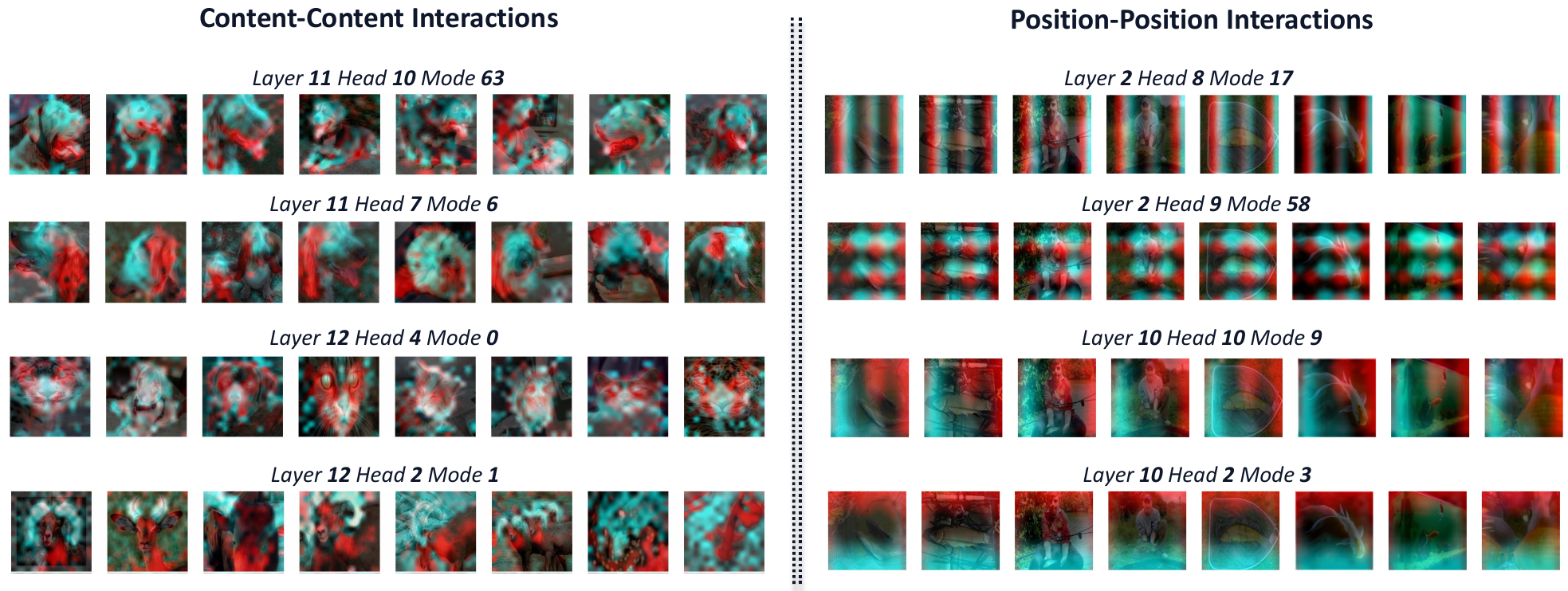}
  \vspace{-5mm}
  \caption{\textbf{Projections of bi-orthogonal modes in DINOv2.}
      The query (red) and key (cyan) singular vectors are projected onto either the content or positional factor, highlighting the image regions that most strongly activate each singular direction.
    }
  \vspace{-3mm}
  \label{fig:main_qualitative_viz}
\end{figure*}

\vspace{-2mm}
\paragraph{Sanity Check: Validating the Decomposition.}
\vspace{-2mm}
Before interrogating attention mechanisms, we first validate that our ANOVA-based factorization genuinely isolates position from content. \cref{fig:sanity_check_position} presents two complementary tests. First, we train linear probes to decode spatial coordinates directly from the activations ($\A$): both ViT and DINOv2 exhibit near-perfect position availability in early layers (left panel), confirming that spatial coordinates are linearly recoverable from the activations themselves. However, a key difference emerges—DINOv2 maintains substantially higher decoding accuracy through intermediate layers (layers 5-8), while ViT's positional availability degrades more rapidly. Both architectures eventually converge to approximately 20\% accuracy in the deepest layers, suggesting that extreme depth compresses positional information regardless of training paradigm. Critically, the content residual $\content$ contains negligible positional information across all layers in both models -- probes trained on content factors alone achieve chance-level decoding accuracy (right panel, performance equivalent to random guessing). This orthogonality by construction, validated empirically, establishes that our decomposition cleanly disentangles the two factors without information leakage. The sustained linear accessibility of position from the block activations in DINOv2’s intermediate layers provides a first hint that self-supervised learning may preserve spatial structure more \textit{explicitly} than supervised training, a hypothesis we will examine geometrically in our analysis of \cref{question:3}. Having confirmed proper factor isolation, we now examine how these factors participate in attention-mediated communication.

\begin{figure}[!b]
    \vspace{-1em}
    \centering
    \includegraphics[width=0.99\linewidth]{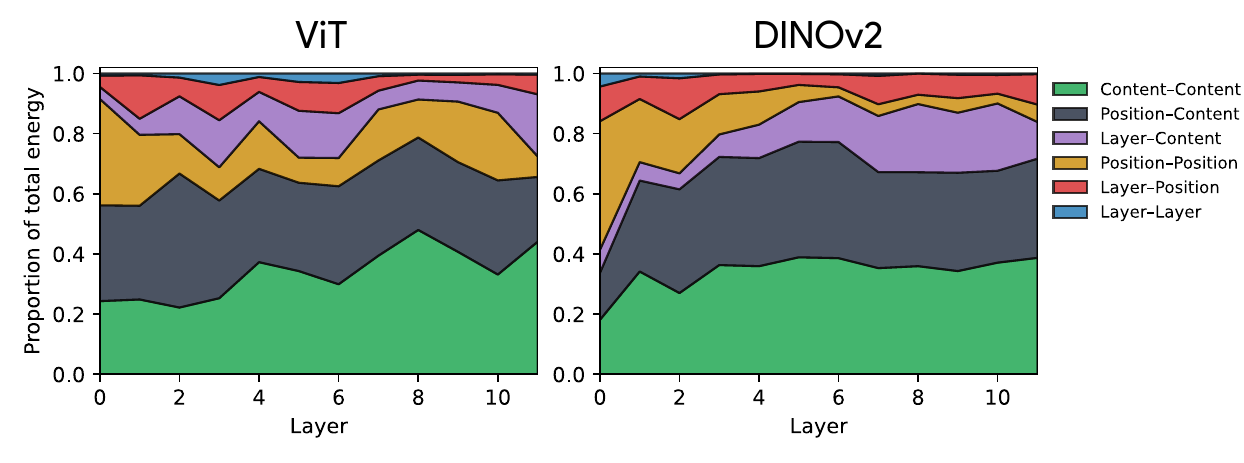}
    \vspace{-3mm}
    \caption{\textbf{Energy Distribution Across Informational Factors.} Layer-wise decomposition of attention energy into contributions from layer effect (global bias), position, content, and their pairwise interactions. Both architectures allocate the majority of energy to content-based interactions (content-content and content-position), confirming that attention-mediated interaction reflects genuine semantic computation rather than positional structure alone. A divergence emerges: DINOv2 dedicates substantially greater energy to content-position coupling across all layers,
    which could be a sign of localization enrichment of semantic content (content gets modulated by positional information).}
    \vspace{-3mm}
    \label{fig:sedimentation}
\end{figure}

\begin{figure*}
    \vspace{-8mm}
    \centering
    \includegraphics[width=0.90\linewidth]{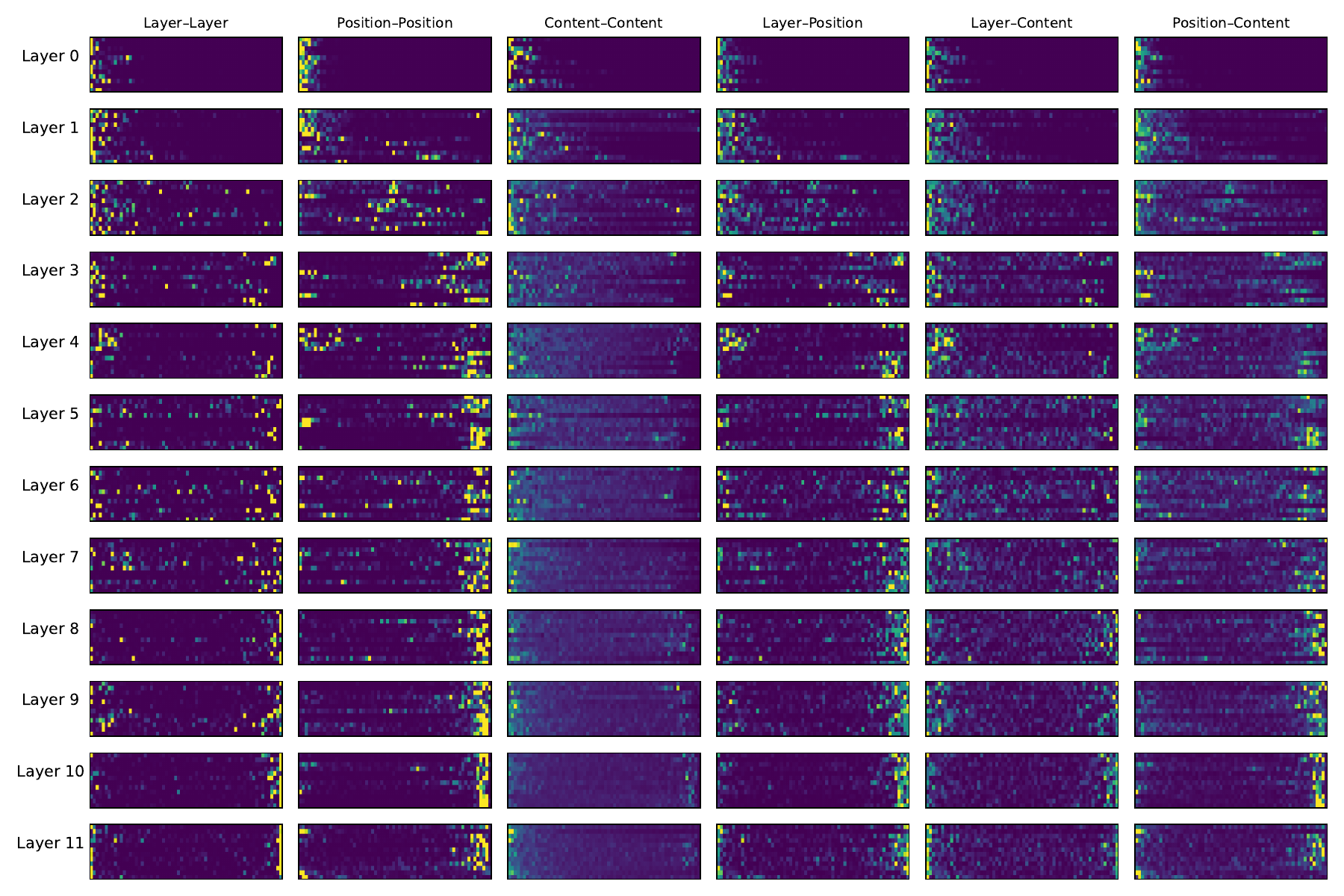}
    \vspace{-3mm}
    \caption{\textbf{Rich Content-Driven Interactions in DINOv2}. Each subplot visualizes the normalized energy for a specific interaction factor (columns) at each layer (rows) for all the decomposed modes. For each layer and interaction, the x-axis represents the bi-orthogonal modes, sorted by singular value from highest (left) to lowest (right) and the y axis represents the head. The color intensity shows the relative contribution of a single mode to a given interaction's total energy (normalized horizontally for each head). The visualization reveals that the low-singular-value modes (right side) also become increasingly dedicated to carrying content-based information in deeper layers.}
    \vspace{-3mm}
    \label{fig:head_mode_energy_DINOv2}
\end{figure*}

\vspace{-2mm}
\paragraph{Characterizing Information Flow.}
\phantomsection
\label{par:infoflow}
\cref{question:1} asks which informational factors—layer effect, position, or content—mediate token communication through attention. We address this by projecting each mode's query and key directions ($\bm{u}_i, \bm{v}_i$) onto the three factors and compute the normalized energy contribution of each bi-orthogonal mode for all unique interactions (see Appendix \ref{app:mode-specialization}). We then aggregate across modes and heads to obtain layer-level profiles (\cref{fig:sedimentation}). Both architectures allocate the majority of attention energy to content-based interactions: content-content (pure semantic exchange) and content-position (localization-aware semantic integration) together dominate the interaction budget, particularly in deeper layers. This finding confirms that contextual integration in later blocks reflects genuine content-driven computation rather than mere positional structure -- the content factor $\content$, which varies across images, accounts for the bulk of query-key alignment (for qualitative examples of these interactions, see \cref{fig:teaser}, \cref{fig:main_qualitative_viz} and Appendix \cref{app:qualitative_visualization}). However, a divergence emerges between training paradigms: DINOv2 dedicates greater energy to content-position than ViT across all layers. At a finer resolution, we observe that this energy is distributed broadly across many modes rather than concentrated in a few dominant ones. As shown in \cref{fig:head_mode_energy_DINOv2}, DINOv2 spreads its content-based interaction energy across a wide range of singular modes, indicating that heads rely on a diverse set of interaction patterns rather than a small set of strong modes. In contrast, the supervised ViT not only allocates less energy to content-based interactions overall, but this energy is also more concentrated in a few dominant modes (see Appendix \cref{fig:app:head_mode_energy_ViT}). 

\begin{figure*}[t]
    \vspace{-10mm}
    \centering
    \includegraphics[width=0.99\linewidth]{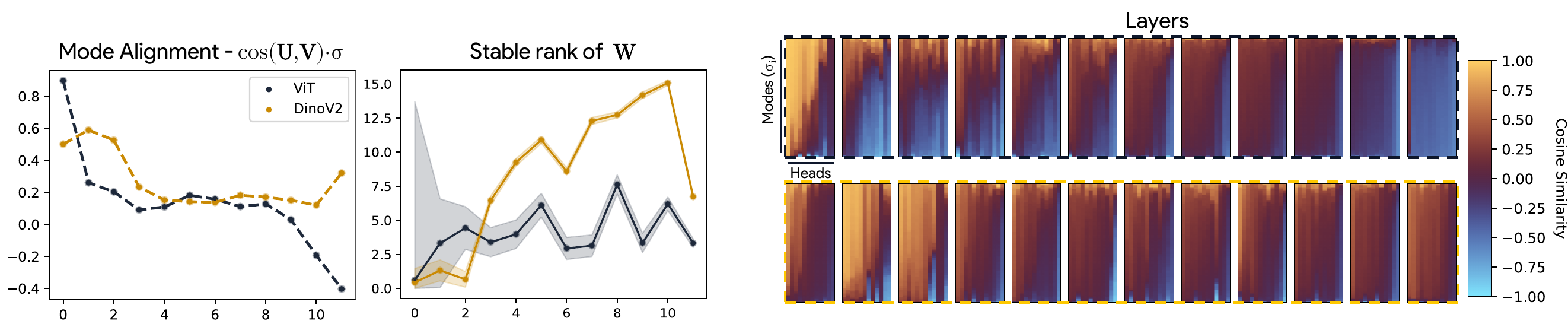}
    \vspace{-2mm}
    \caption{\textbf{Spectral Structure of Query-Key Interactions.} \textit{Left:} Mode alignment $\cos(\bm{u}_i, \bm{v}_i) \cdot \sigma_i$ across layers. Both architectures transition from high alignment (symmetric modes) in early layers to near-orthogonality at depth, consistent with prior observations~\cite{pan2024dissecting}. A subtle but important difference emerges: ViT exhibits negative cosine values (anti-aligned modes), while DINOv2 approaches contextual grouping behavior with modes clustering near orthogonality rather than anti-alignment. \textit{Middle:} Stable rank of the interaction matrix $\bm{W}$ reveals that DINOv2 maintains substantially higher effective dimensionality across all layers, indicating that its attention operates through a richer subspace with more distributed communication channels. \textit{Right:} Mode-specific alignment distributions refine the aggregate view. ViT concentrates mass in negative-alignment modes, whereas DINOv2 exhibits a more balanced distribution. Critically, within each architecture, modes differentiate into specialized regimes -- some highly aligned (symmetric query-key relationships), others orthogonal or anti-aligned -- establishing mode-level functional diversity beyond head-level aggregates.}
    \vspace{-3mm}
    \label{fig:modes_details}
\end{figure*}

To understand whether this energy distribution reflects distributed computation or modular specialization, we next examine the spectral structure of the interaction matrix $\bm{W} = \bm{W}_Q\bm{W}_K^\top$ itself. ~\cref{fig:modes_details} (left panels) reveals that DINOv2 maintains substantially higher stable rank than ViT across layers, indicating that its attention operates through a richer, more distributed set of communication modes rather than concentrating interaction on a few dominant directions. This higher rank is a direct consequence of a flatter singular value spectrum; DINOv2 sustains a greater number of modes with high singular values compared to the supervised ViT, whose spectrum decays more rapidly (see Appendix \cref{fig:app:details_singular_value_w} for a detailed layer-by-layer view). The alignment between query and key modes, quantified as $\cos(\bm{u}_i, \bm{v}_i) \cdot \sigma_i$, further distinguishes the two models (\cref{fig:modes_details}; right panel): ViT exhibits high alignment (modes are nearly symmetric), whereas DINOv2 displays lower alignment, particularly in intermediate layers, signaling asymmetric query-key specialization. This spectral signature: higher rank, asymmetric modes, and high content-position energy, constitutes a quantitative fingerprint of self-supervised attention. Yet the existence of multiple modes raises a natural question: do these modes serve distinct functional roles, or do they redundantly encode the same information?

\vspace{-2mm}
\paragraph{Functional Specialization: Mode Purity and Head Differentiation.} 
\vspace{-2mm}

\begin{figure}[t]
    \vspace{-1mm}
    \centering
    \includegraphics[width=0.99\linewidth]{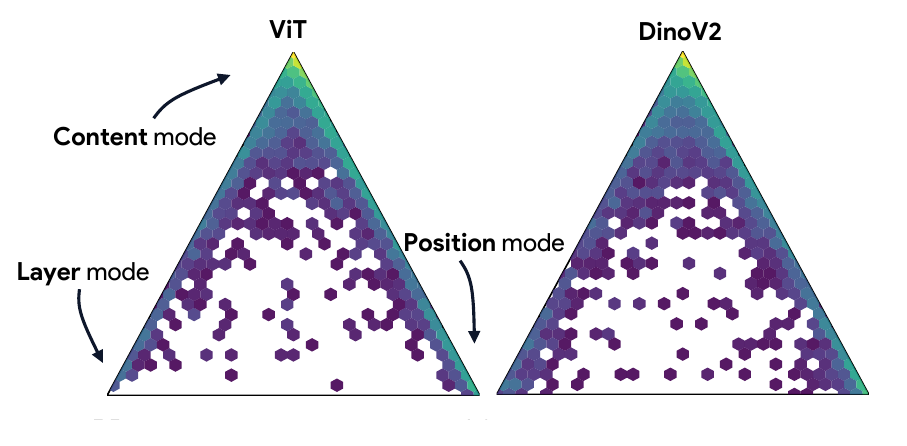}
    \vspace{-3mm}
    \caption{\textbf{Functional specialization of Attention Modes.} Ternary plots showing the barycentric coordinates of each bi-orthogonal mode across all the heads and layers in ViT (left) and DINOv2 (right). Each point reflects the relative contribution of layer-, position-, and content-related interactions for a mode. Modes cluster near the vertices rather than dispersing uniformly across the simplex, revealing strong functional specialization for both the models, with tighter clustering in DINOv2, indicating more pronounced specialization under self-supervised training.}
    \vspace{-3mm}
    \label{fig:specialization}
\end{figure}

\cref{question:2} concerns whether attention heads and their constituent singular modes exhibit functional specialization into distinct informational operators. We address this by projecting each mode’s query and key directions $(\bm{u}_i,\bm{v}_i)$ onto the three informational factors -- layer, position,
and content and computing the normalized interaction energies for the six undirected factor pairs (see Appendix \cref{app:mode-specialization}). We then group these six interactions into three families (layer, position, and content) and compute barycentric coordinates that indicate the relative contribution of layer-, position-, and content-related interactions for each mode. Visualizing the distribution of modes in these coordinates using ternary plots (\cref{fig:specialization}) — where the vertices correspond to layer-dominated, position-dominated, and content-dominated operators, and points along edges reflect mixed operators (e.g., content–position)—reveals strong mode purity: individual modes cluster near specific vertices rather than dispersing uniformly across the simplex.

This indicates that most modes implement specialized informational operations (Appendix \cref{app:qualitative_visualization}): some mediate pure semantic exchange (content-content), others mainly track spatial relationships (position-position), and still others integrate semantics with spatial context (content-with-position). Crucially, this specialization holds both at the mode level (individual singular vectors within a head) and at the head level (when aggregating energy across all modes of a head). Across both ViT and DINOv2, heads differentiate into three functional classes, with DINOv2 exhibiting slightly greater mode purity and more pronounced differentiation in intermediate layers (Appendix ~\cref{fig:app:mode_specialization_points} and Appendix ~\cref{fig:app:mode_triangle_full_density} provide per-layer plots). This functional modularity implies that per-head mechanistic analyses are not fundamentally incomplete; \method~decomposition of heads finds pure computational primitives that can be meaningfully composed. The consistency of specialization across architectures suggests it is an emergent property of multi-head self-attention rather than a training-specific artifact, though the degree of purity varies with supervision. Having established that attention mechanisms decompose into functionally interpretable operators, we now ask: what representational properties enable DINOv2's superior holistic shape processing, and how do these properties manifest in the geometry of learned representations?

\begin{figure*}[t]
    \vspace{-8mm}
    \centering
    \includegraphics[width=0.99\linewidth]{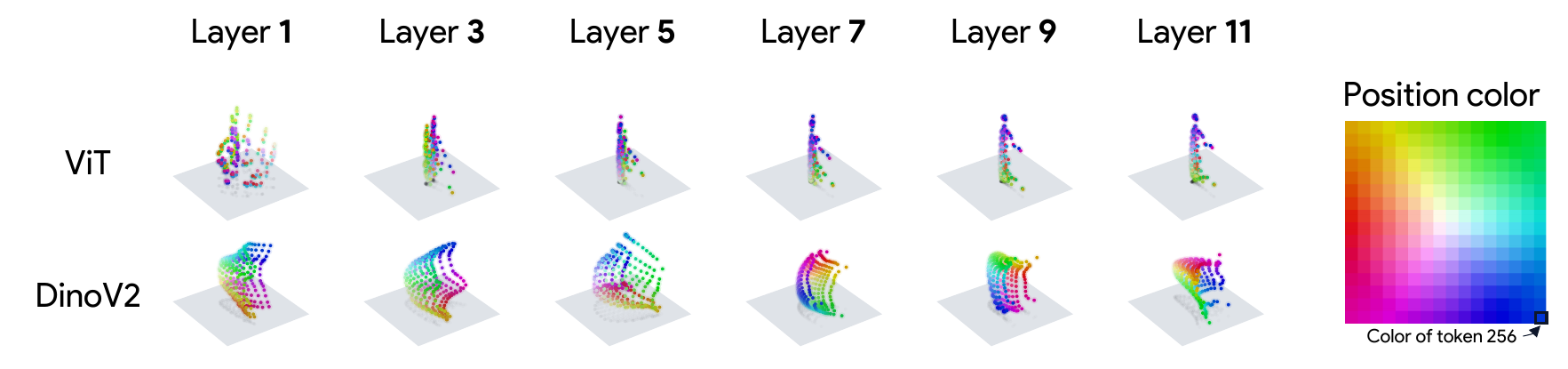}
    \caption{\textbf{Geometric Preservation of Positional Structure.} PCA visualization of positional factors $\position$ across layers, with each token's spatial location encoded by color. The first three principal components are rendered as RGB-encoded 3D point clouds. \textit{Top (ViT):} Supervised training progressively collapses the spatial manifold—by layer 5, the 2D grid degenerates into a quasi-1D structure with tokens condensing along a single principal axis, obliterating neighborhood relationships. \textit{Bottom (DINOv2):} Self-supervised training preserves the 2D spatial grid topology throughout all layers. Tokens retain their neighborhood structure, and the manifold remains approximately planar (a 2D sheet embedded in 3D space) even at depth.
    }
    \label{fig:pca_position}
    \vspace{-3mm}
\end{figure*}

\begin{figure}[t]
    \vspace{-2mm}
    \centering
    \includegraphics[width=0.99\linewidth]{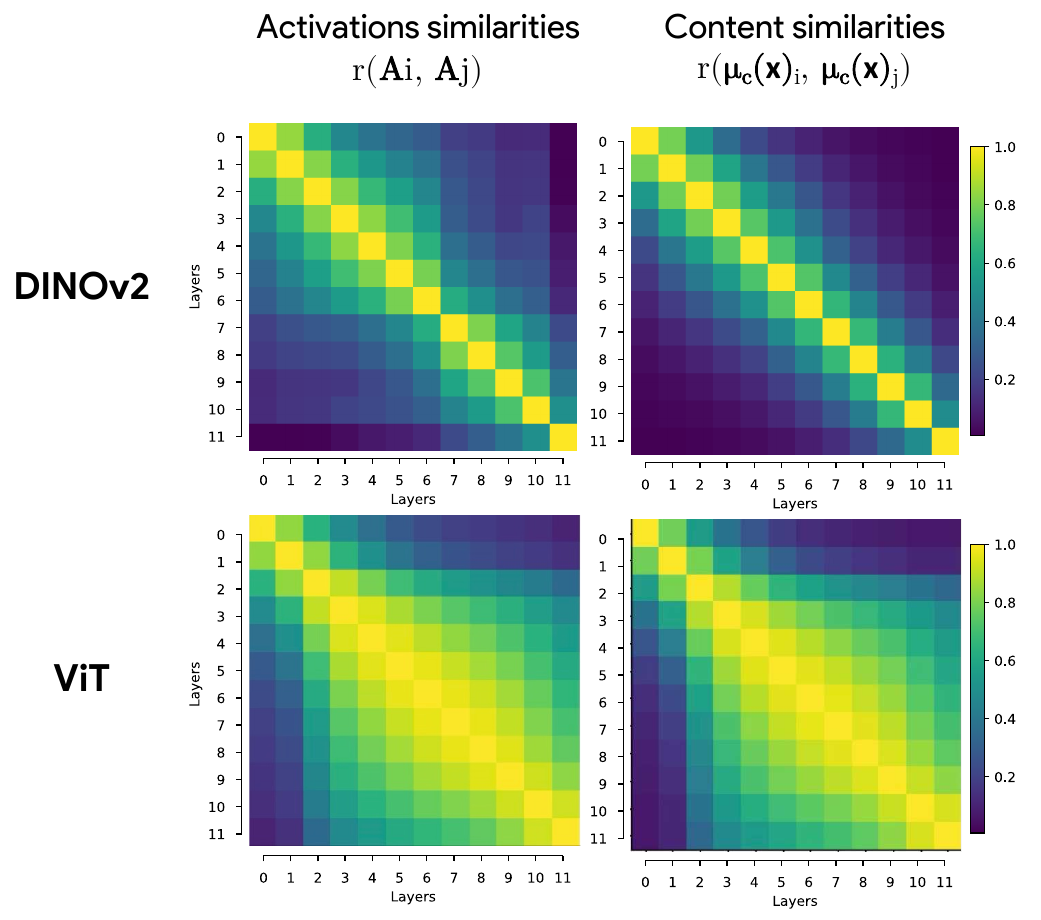}
    \caption{\textbf{Content Enrichment with Depth.} Pairwise Pearson correlations across layers for full activations ($A_i$) and content factors ($\mu_c(x)_i$). \textit{DINOv2 (top)} shows a gradual transformation of content across depth, revealed once positional signals are removed. In \textit{ViT (bottom)}, content and activations show similar trajectories, indicating weaker positional influence and limited semantic evolution.}
    \label{fig:content_flow}
    \vspace{-5mm}
\end{figure}

\paragraph{Dual Preservation: Positional Structure and Semantic Enrichment as the Signature of Holistic Processing.}
~\cref{question:3} asks what structural properties of intermediate-layer representations distinguish self-supervised from supervised models and enable holistic shape processing~\cite{geirhos2018imagenet,doshi2025visual}. To probe the geometry of positional representations, we apply PCA to the positional factor $\position$ across all 196 and 256 spatial tokens and visualize the first three principal components as RGB-encoded 3D point clouds (~\cref{fig:pca_position}). A divergence emerges: DINOv2 preserves the 2D spatial grid topology throughout all layers—tokens retain their neighborhood structure, and the manifold remains approximately planar (2D sheet embedded in 3D space) even at depth. In contrast, ViT progressively collapses the spatial manifold into a quasi-1D structure by layer 5, with tokens condensing along a single principal axis and obliterating local spatial relationships (Appendix \cref{fig:app:details_position_ViT} and \cref{fig:app:details_position_DINOv2} provide rotated views across all layers). While linear probes confirm that positional information remains decodable in both models, the collapse of the positional subspace dimensionality in ViT suggests its spatial coordinate system becomes far less expressive in mid-to-later layers. This positional preservation in DINOv2 directly addresses the first half of Question~\ref{question:3}: self-supervised models maintain ``explicit'' spatial structure where supervised models do not. Yet preservation alone does not explain holistic processing—the model must also enrich semantic content.

Next, we examine how semantic content evolves across the network. We analyze this by computing the Pearson correlation of content factors  and the full activations, of all tokens, for each pair of layers ($r(\contentimagei, \contentimagej)$ and $r(\A_i, \A_j)$) in ~\cref{fig:content_flow}. While the full activations show a higher degree of similarity across all layers, with similarity decaying slowly from the diagonal, the similarity marix for the content factor exhibits a smoother enrichment trajectory, with incremental refinement through the intermediate layers. The correlation between early- and late-layer's content factor is significantly lower than for the full activations. This indicates that much of the apparent similarity in the full activations is an artifact of the persistent positional code. Once the  positional signal is removed, it is clear that the underlying semantic content is undergoing a transformation as it progresses through each layer the network. 

In contrast, the ViT similarity structure reveals a different pattern. The layer–layer correlations for the content factors closely mirror those of the full activations, despite the content factors being position-free. This indicates that positional information plays a far smaller role in shaping ViT’s internal geometry, consistent with the early collapse of the positional scaffold we observed earlier. More importantly, the content factors themselves show only limited evolution across depth: early and late layers remain highly correlated. This suggests that, rather than progressively enriching semantic content through contextual interactions, the ViT primarily performs more local refinement (e.g., refining an “eye” into a slightly sharper “eye”) rather than integrating it into a broader semantic configuration (e.g., placing the eye within the context of a face). DINOv2, by contrast, shows a steady transformation of its content factors across layers, indicative of contextually enriched representations: token meaning is updated by accumulating information from other parts of the object (see example for semantic interactions in Appendix \cref{fig:app:qualitative_position_position_interaction_blockindex_1}). This rapid evolution of semantic content in DINOv2, is the second key component of the dual preservation. We believe that the dual preservation -- maintaining a stable spatial topology while enabling the evolution of semantic content -- constitutes the computational motif underlying holistic shape processing. 

\vspace{-1mm}
\section{Discussion}
\vspace{-1mm}
We introduced Bi-orthogonal Factor Decomposition (\method{}), a framework that jointly factorizes vision transformer activations into positional and content components and spectrally decomposes their attention interactions. This coupling exposes what type of information flows through self-attention and how it is organized across heads and modes. When applied to supervised and self-supervised ViTs, \method{} reveals a consistent computational structure within self-attention. First, heads and their singular modes exhibit clear functional specialization: rather than mixing informational factors, each channel reliably operates as a position, content, or mixed content–position operator. Second, both architectures allocate most of their attention energy to content–dominated interactions, indicating that semantic features, not positional offsets, drive the majority of token-to-token communication. Third, this pattern is amplified in DINOv2: content-dominated interactions are stronger and distributed across a larger set of modes, reflecting a richer and more distributed computational motif than in supervised ViTs. Finally, self-supervised models maintain a well-structured positional scaffold while their content representations progressively enrich across depth, enabling coordinated spatio-semantic integration that is largely absent in supervised counterparts, hence offering a mechanistic explanation for their superior holistic shape processing.

{
    \small
    \bibliographystyle{ieeenat_fullname}
    \bibliography{main}
}
\clearpage

\appendix

\clearpage
\section{Qualitative Visualization of Interactions from Different Information Factors}
\label{app:qualitative_visualization}

\begin{figure*}[!t]
  \centering
  \includegraphics[width=\linewidth]{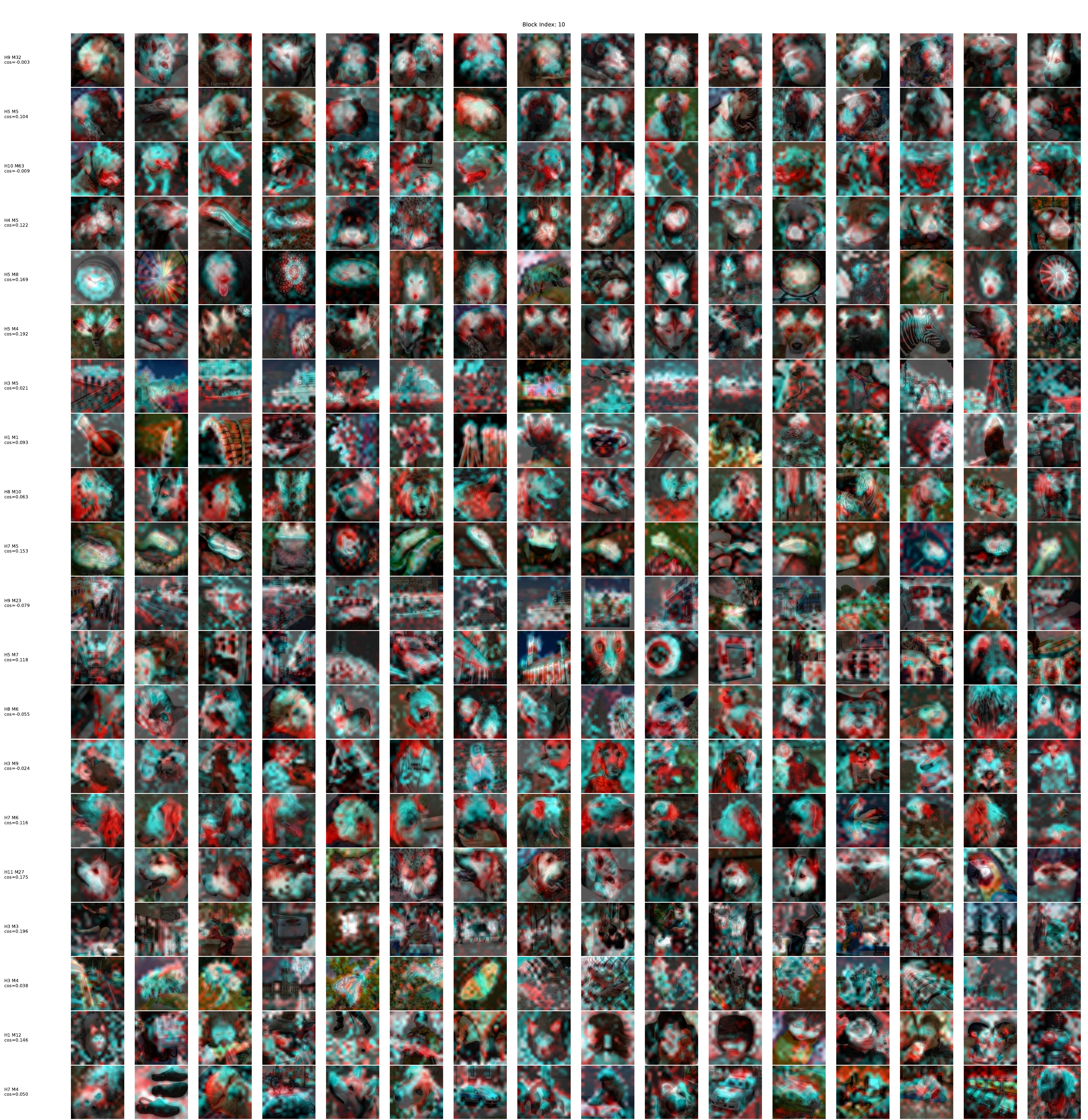}
  \caption{\textbf{Content-Content Interactions in DINOv2 Block Index 10.}
    }
  \label{fig:app:qualitative_content_content_interaction_blockindex_10}
\end{figure*}

\begin{figure*}[!t]
  \centering
  \includegraphics[width=\linewidth]{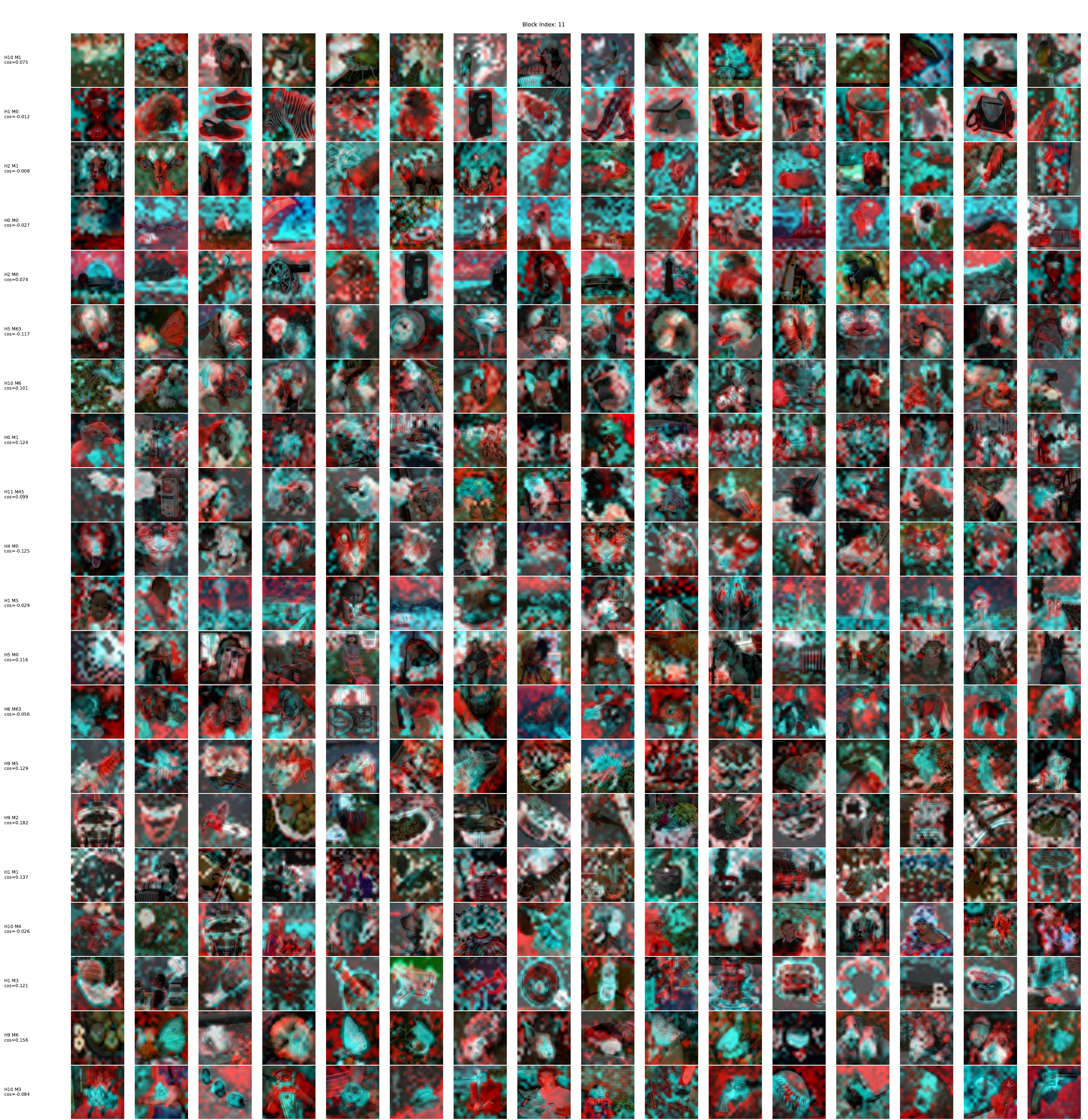}
  \caption{\textbf{Content-Content Interactions in DINOv2 Block Index 11.}
    }
  \label{fig:app:qualitative_content_content_interaction_blockindex_11}
\end{figure*}

\begin{figure*}[!t]
  \centering
  \includegraphics[width=\linewidth]{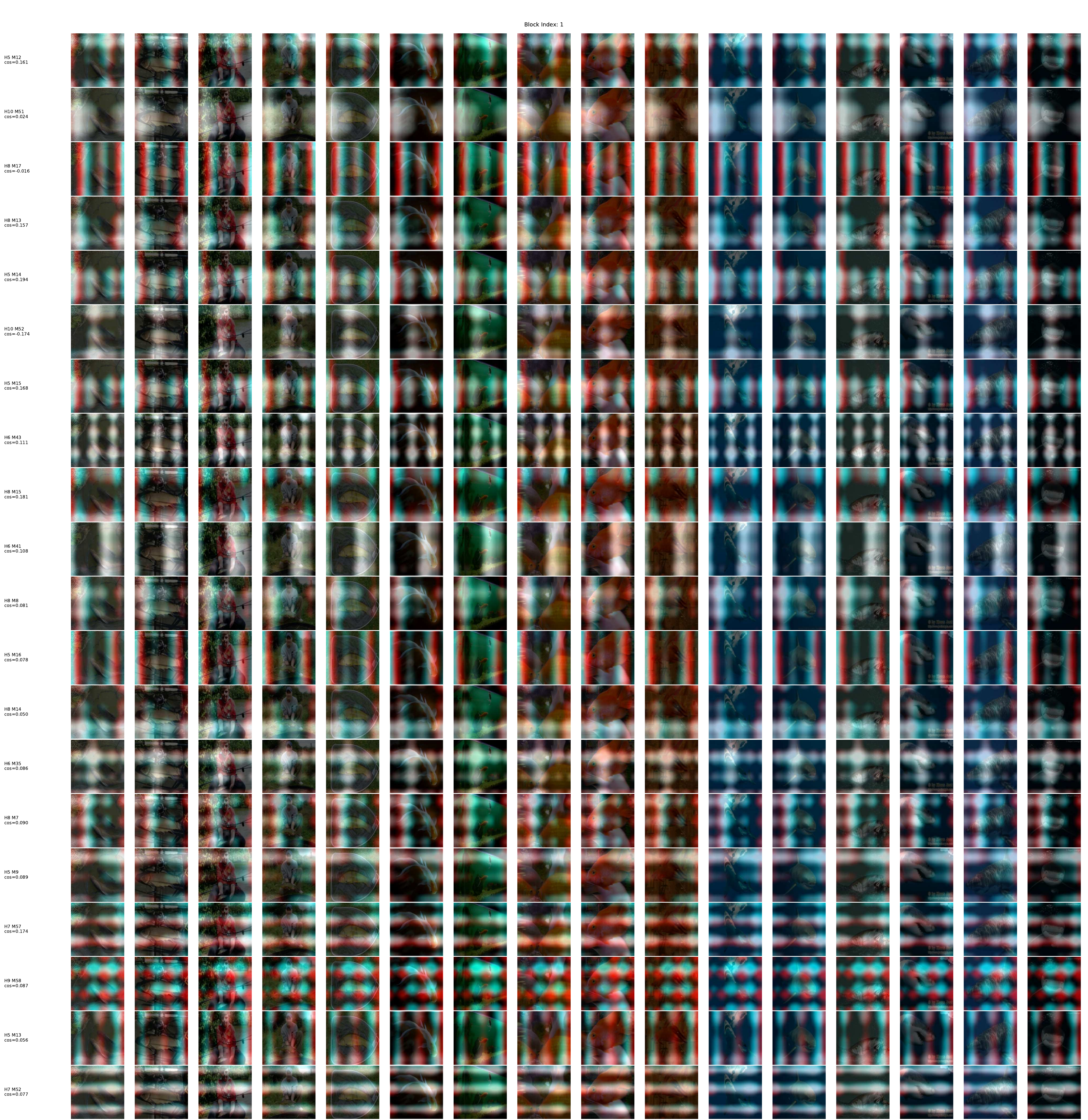}
  \caption{\textbf{Position-Position Interactions in DINOv2 Block Index 1.}
    }
  \label{fig:app:qualitative_position_position_interaction_blockindex_1}
\end{figure*}

\begin{figure*}[!t]
  \centering
  \includegraphics[width=\linewidth]{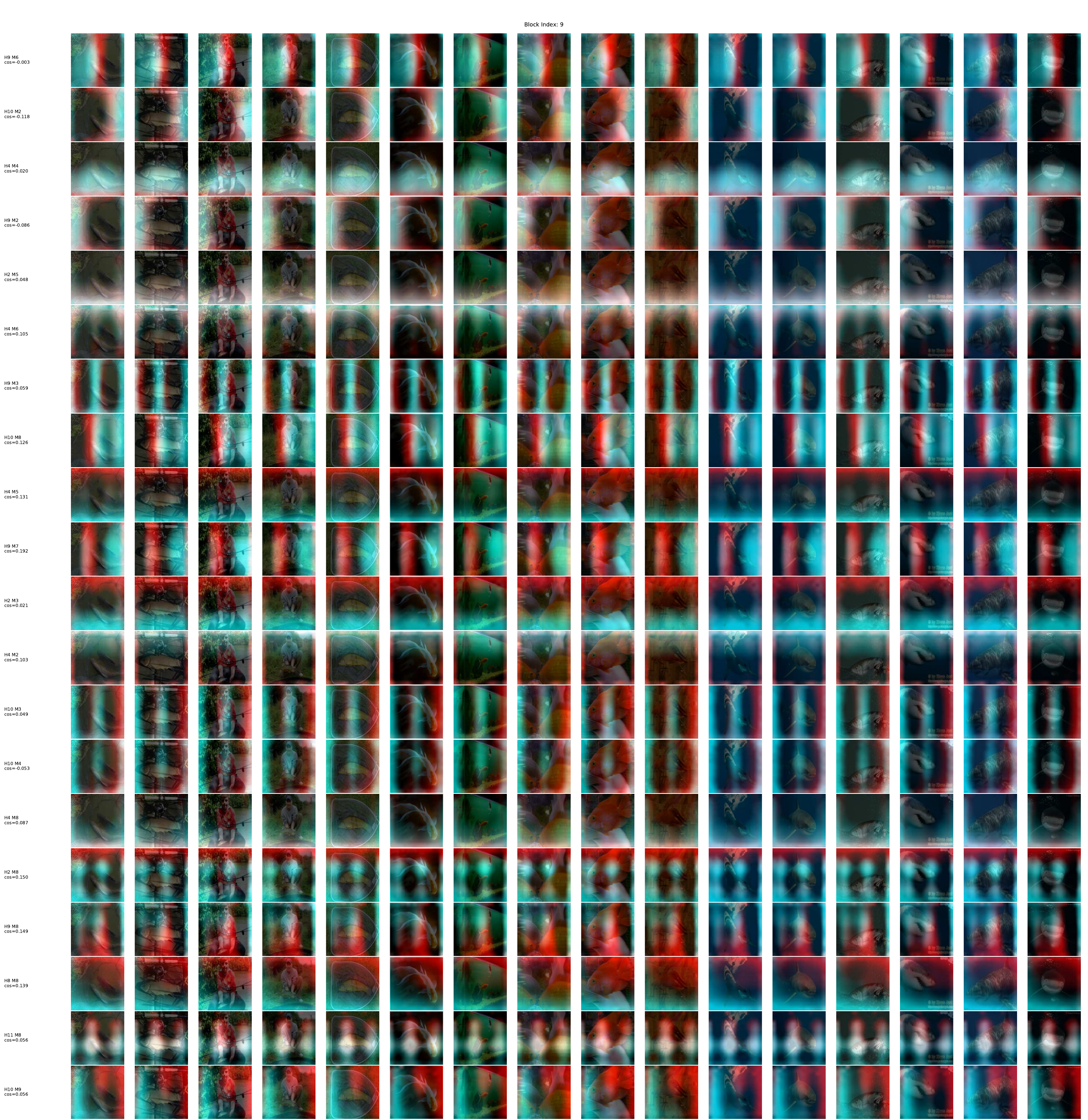}
  \caption{\textbf{Position-Position Interactions in DINOv2 Block Index 9.}
    }
  \label{fig:app:qualitative_position_position_interaction_blockindex_9}
\end{figure*}

\begin{figure*}[!t]
  \centering
  \includegraphics[width=\linewidth]{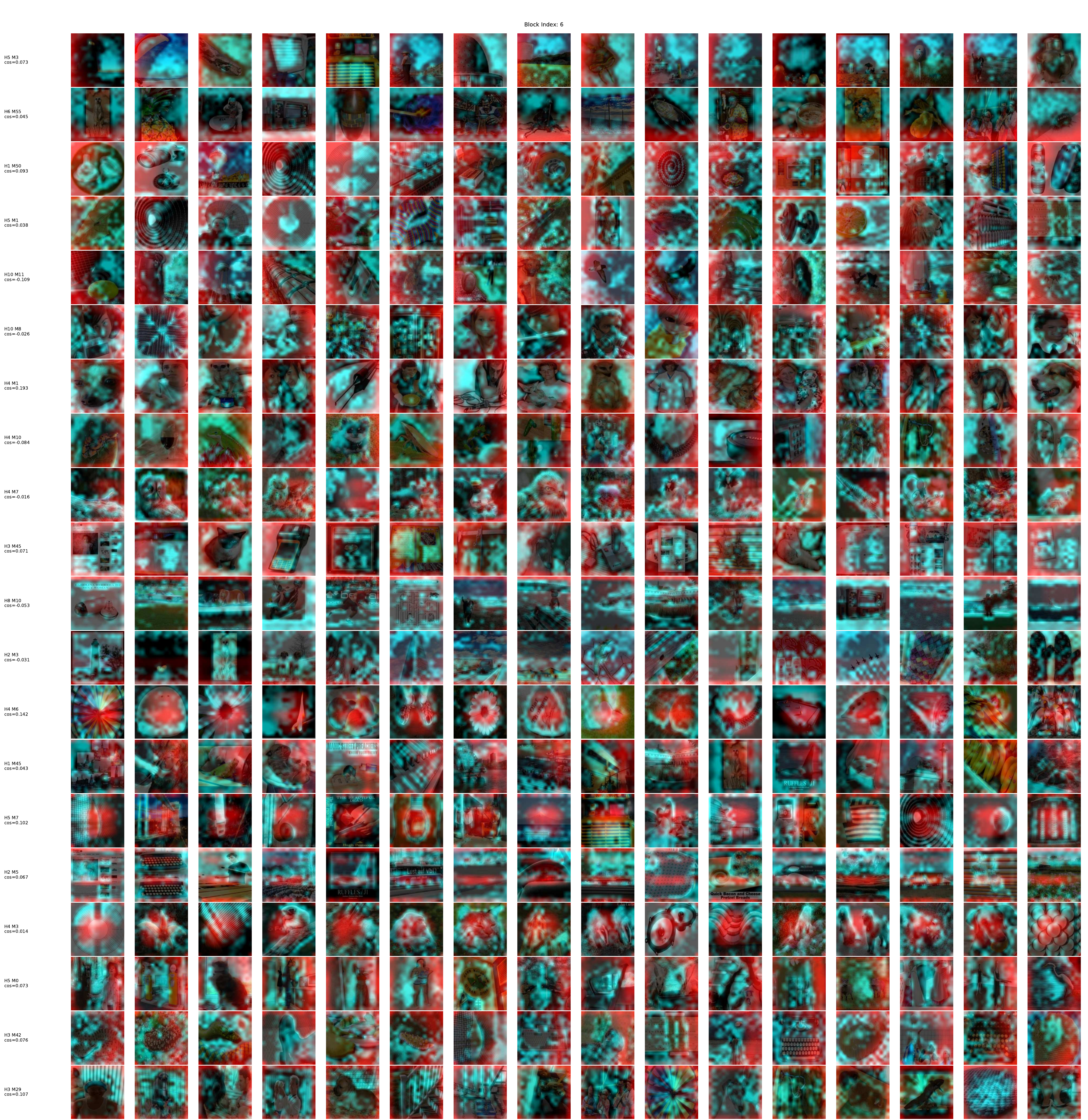}
  \caption{\textbf{Position-Content Interactions in DINOv2 Block Index 6.}
    }
  \label{fig:app:qualitative_position_content_interaction_blockindex_6}
\end{figure*}

\begin{figure*}[!t]
  \centering
  \includegraphics[width=\linewidth]{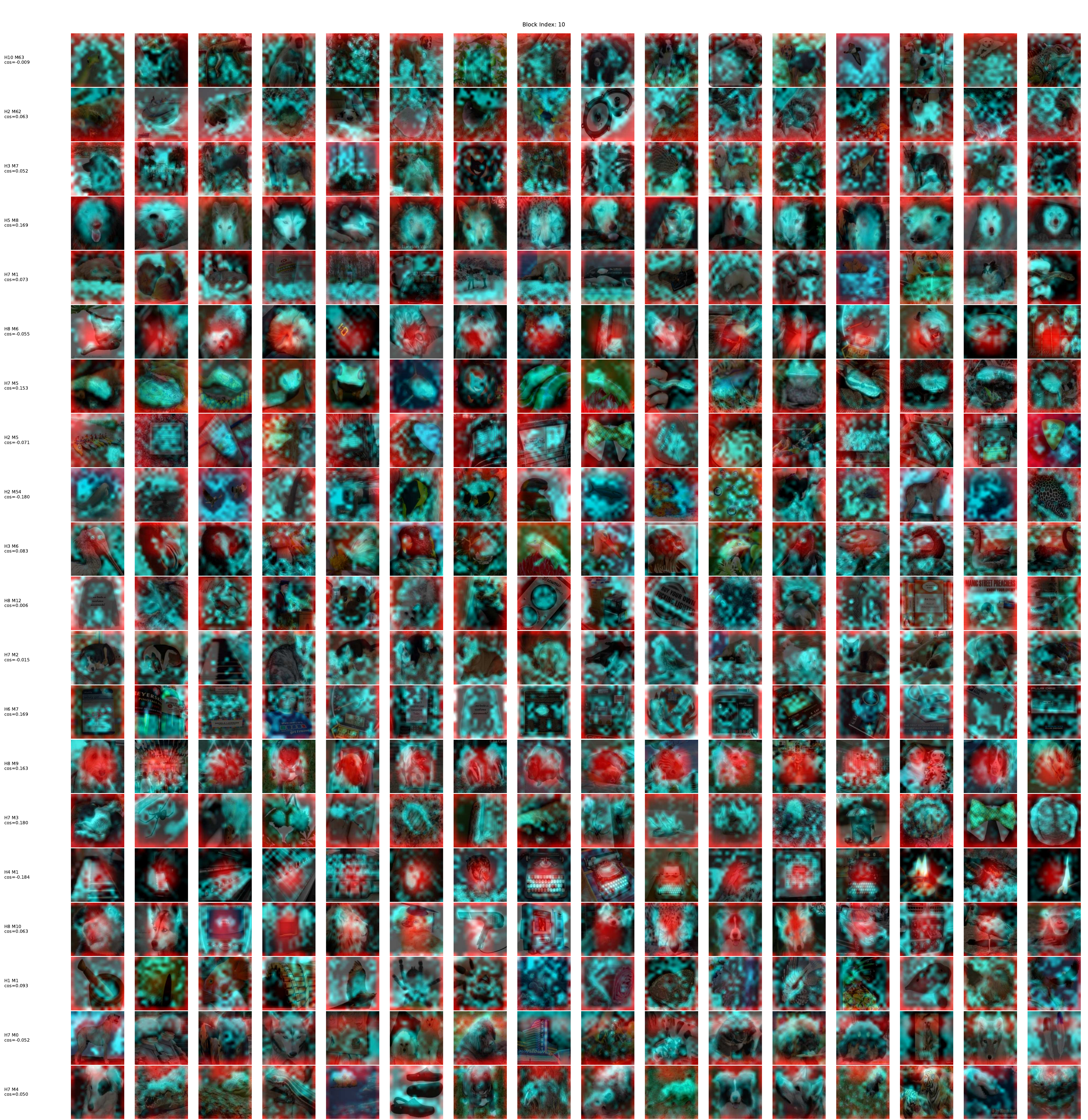}
  \caption{\textbf{Position-Content Interactions in DINOv2 Block Index 10.}
    }
  \label{fig:app:qualitative_position_content_interaction_blockindex_10}
\end{figure*}

To complement the quantitative analyses in the main paper, we present qualitative visualizations of  three key interactions extracted by BFD: content–content, content–position, and position–position. For each type, we show representative bi-orthogonal modes whose query–key directions $(\bm{u}_i,\bm{v}_i)$ are activated by that specific interaction. Each visualization overlays the mode’s spatial pattern onto the activating images, providing intuition about what kind of signal the mode reads from the query and the key token.

\paragraph{Content–Content (Purely Semantic Interactions).}
These modes activate on meaningful interactions between object parts -— faces, limbs, border edges -—indicating semantic affinity between regions. Because both query and key directions respond to content structure, they highlight \emph{what} the object is, not \emph{where} it is. Qualitatively, they reveal part-to-part or part-to-whole correspondences (e.g., eye-to-head or ear-to-snout), demonstrating how semantic information is exchanged between tokens.

\paragraph{Position–Position (Purely Spatial Interactions).}
These modes activate on geometric structure rather than semantics. They highlight global spatial patterns such as left-to-right sweeps, top-to-bottom gradients, or Fourier-like waveforms. These interactions allow the modes to keep a track of how positional information flows through tokens.

\paragraph{Content–Position (Localization-Aware Semantic Interactions).}
These modes activate on semantic regions in a way that depends on spatial context. One side (query or key) aligns with semantic content while the other aligns with positional structure. As a result, they highlight meaningful regions that are modulated by where they occur in the image -- e.g., activating on specific object parts based on stereotyped spatial queries. These patterns illustrate how semantic features directly interact with positional cues.

\paragraph{}Together, these qualitative examples show how different interaction types manifest at the mode level—semantic exchange, spatial exchange, and spatially conditioned semantic exchange—and provide intuition for the functional specialization results reported in the main text. 

\FloatBarrier
\clearpage

\clearpage
\section{Linear Probe Evaluation for Factorization}
\label{app:linear-probe}

To verify that the factorization cleanly separates positional and
content structure, we train token-level linear probes to decode the
spatial coordinate of each patch token from (i) the unfactorized block
activations and (ii) the content factor $\mu_{c}$.

\paragraph{Data preparation.}
For each model and each layer, we extract either the block activations
$A_{\ell}(\x) \in \mathbb{R}^{T \times D}$ or the content factor
$\mu_{c}(\x) \in \mathbb{R}^{T \times D}$ from $5{,}000$ ImageNet
images, where $T$ is the number of patch tokens and
$D$ is the embedding dimension. We treat every patch token as an individual training example. With 5{,}000 images and 256 patch tokens per image, this yields $ N = 5000 \times 256 = 1.28M$ token examples per probe. Each token has a discrete position label $\in \{1,\dots,256\}$ corresponding to its grid location. We collect the token representations into a matrix $\mathbf{H}$ and their position labels into a vector $\mathbf{y}$:

\begin{equation*}
\begin{aligned}
    \mathbf{H} &\in \mathbb{R}^{N \times D}, \\
    \mathbf{y} &\in \{1,\dots,256\}^{N}, \\
    N &= 5000 \times 256 = 1{,}280{,}000.
\end{aligned}
\end{equation*}

\paragraph{Probe architecture and optimization.}
A multinomial logistic regression classifier is trained using cross-entropy loss.  
We use the Adam optimizer (learning rate $10^{-2}$, batch size
8{,}192, 20 epochs), implemented in a GPU-optimized batched training
loop. No nonlinearities, regularization, or data augmentation are used. After training, the probe is evaluated on all tokens. Accuracy reflects the fraction of tokens for which the predicted
coordinate matches the ground-truth position. High accuracy when decoding from the unfactorized block activations
confirms that spatial topology remains linearly accessible in the raw
representation, whereas chance-level accuracy from the content factor
$\mu_{c}$ verifies that positional information is successfully removed
by the factorization.

\FloatBarrier

\section{Interaction Maps and Mode Specialization}
\label{app:mode-specialization}

\newcommand{\uvec}{\mathbf{u}}
\newcommand{\vvec}{\mathbf{v}}

\newcommand{\U}{\mathbf{U}}

\newcommand{\Sig}{\boldsymbol{\Sigma}}
\newcommand{\PhiL}{\boldsymbol{\Phi}_{\!L}}
\newcommand{\PhiP}{\boldsymbol{\Phi}_{\!P}}
\newcommand{\PhiC}{\boldsymbol{\Phi}_{\!C}}

\paragraph{Factor-projected mode energies (per image).}

For head $h$ with SVD $\W=\U\Sig\V^\top$ and mode $i$ (columns
$\uvec_{i},\vvec_{i}$; singular value $\sigma_{i}$), we compute the
factor-projected query/key codes ($\mu^{Q}, \mu^{K} in\{\mu_L,\mu_P,\mu_C\}$) for image $\x$:
\begin{align*}
  \z^{Q}_{\cdot,i}(\x) &= \mu^{Q}(\x)\,\uvec_{h,i}\in\mathbb{R}^{d},\\
  \z^{K}_{\cdot,i}(\x) &= \mu^{K}(\x)\,\vvec_{h,i}\in\mathbb{R}^{d}
\end{align*}

The factor-projected energy for mode $i$ in image $\x$ is then:
\begin{equation*}
  \energy^{(i)}_{\cdot}(\x)
  \;=\;
  \big\|\, \bm{z}^{Q}_{\cdot,i}(\x)\; \sigma_i\; \bm{z}^{K}_{\cdot,i}(\x) \,\big\|_2^2.
\end{equation*}

\paragraph{Mode Normalization.}

After computing the per-image mode energies $\energy^{(i)}_{\cdot}(\x)$,
we normalize across all modes within the same head and interaction type,
and then take the expectation across images:

\begin{equation*}
    \bar{\energy}^{(i)}_{\cdot}
    = \mathbb{E}_{\x}
      \left(
        \frac{
          \energy^{(i)}_{\cdot}(\x)
        }{
          \sum_{j} \energy^{(j)}_{\cdot}(\x)
        }
      \right),
    \qquad
    \sum_{i} \bar{\energy}^{(i)}_{\cdot} = 1.
\end{equation*}

\paragraph{From nine to six interactions (symmetrization).}
We symmetrize directional pairs:
\begin{align*}
  \overline{\mathcal{E}}^{(i)}_{P\!-\!C}
  &= \tfrac{1}{2}\!\left(\overline{\mathcal{E}}^{(i)}_{P\!-\!C}+\overline{\mathcal{E}}^{(i)}_{C\!-\!P}\right),\\
  \overline{\mathcal{E}}^{(i)}_{L\!-\!C}
  &= \tfrac{1}{2}\!\left(\overline{\mathcal{E}}^{(i)}_{L\!-\!C}+\overline{\mathcal{E}}^{(i)}_{C\!-\!L}\right),\\
  \overline{\mathcal{E}}^{(i)}_{L\!-\!P}
  &= \tfrac{1}{2}\!\left(\overline{\mathcal{E}}^{(i)}_{L\!-\!P}+\overline{\mathcal{E}}^{(i)}_{P\!-\!L}\right),
\end{align*}
yielding six undirected interactions
$\{L\!-\!L,\,P\!-\!P,\,C\!-\!C,\,L\!-\!P,\,L\!-\!C,\,P\!-\!C\}$.

\paragraph{Interaction Maps}

\begin{figure*}
    \centering
    \includegraphics[width=0.99\linewidth]{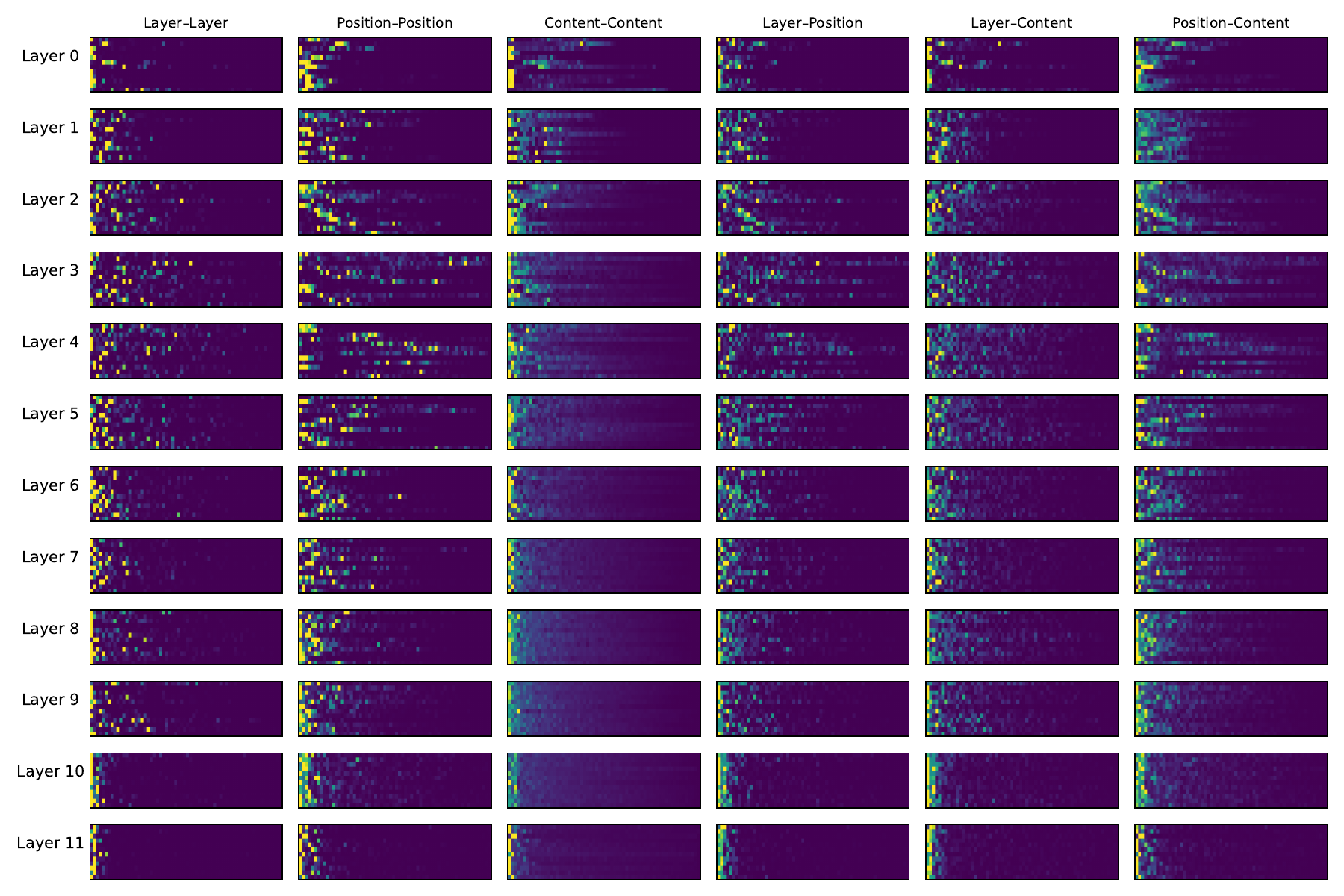}
    \caption{\textbf{Normalized Energy per Mode and Head for ViT}. Each subplot visualizes the normalized energy for a specific interaction factor (columns) at each layer (rows). For each layer and interaction, the x-axis represents the bi-orthogonal modes, sorted by singular value from highest (left) to lowest (right) and the y axis represents the head. The color intensity shows the relative contribution of a single mode to a given interaction's total energy (normalized horizontally for each head). The visualization reveals that the 
    energy is concentrated in the select few top singular modes.}
    \vspace{-5mm}
    \label{fig:app:head_mode_energy_ViT}
\end{figure*}

For each transformer layer $\ell$ and each of the six symmetrized interaction types, we compute their normalized mode
energies $\overline{\mathcal{E}}^{(i)}_{L\!-\!L}, \overline{\mathcal{E}}^{(i)}_{P\!-\!P}, \overline{\mathcal{E}}^{(i)}_{C\!-\!C}, \overline{\mathcal{E}}^{(i)}_{L\!-\!P}, \overline{\mathcal{E}}^{(i)}_{L\!-\!C}, \overline{\mathcal{E}}^{(i)}_{P\!-\!C}$. These energies are arranged into
head--by--mode matrices, with heads (12 per layer) along the rows and
singular modes (64 per head) along the columns (Appendix
\cref{fig:app:head_mode_energy_ViT} and \cref{fig:head_mode_energy_DINOv2}). Brighter values
indicate modes that contribute more strongly to a given interaction after
within-interaction normalization.

These maps reveal two consistent trends. First, fine-grained specialization
emerges at the level of individual modes rather than at the level of
whole heads: even within a single head, different modes selectively
support different informational interactions. Second, the
distribution of energy across modes differs systematically between ViT
and DINOv2. In the supervised ViT, interaction energy is concentrated
into a small subset of top singular modes, producing visibly sharper
matrices. In contrast, DINOv2 exhibits a more distributed pattern, with
content-driven and position-driven interactions expressed across a
broader range of modes. This provides a layer-wise view of the richer
mode spectrum and higher stable rank highlighted in the main text.

\paragraph{Ternary plots.}
To visualize mode specialization, we group the six undirected interaction energies for each mode $i$ into three informational families: a \emph{layer} family (all interactions involving the layer factor), a
\emph{position} family (interactions involving the position factor), and a \emph{content} family (interactions involving the content factor). Mixed interactions (such as position–content or layer–content) contribute to both relevant families. We then normalize the three family scores to obtain barycentric
coordinates for each mode. Each singular mode $i$ is plotted at $\big(\hat{S}^{(i)}_{L},\hat{S}^{(i)}_{P},\hat{S}^{(i)}_{C}\big)$ in the ternary simplex. Points near the content vertex are dominated by content-only (resp.\ position-only, layer-only) interactions.   Points along the content–position edge indicate genuine co-activation of these families (localization-aware semantic integration).

\begin{figure}[t]
    \centering
    \includegraphics[width=0.99\linewidth]{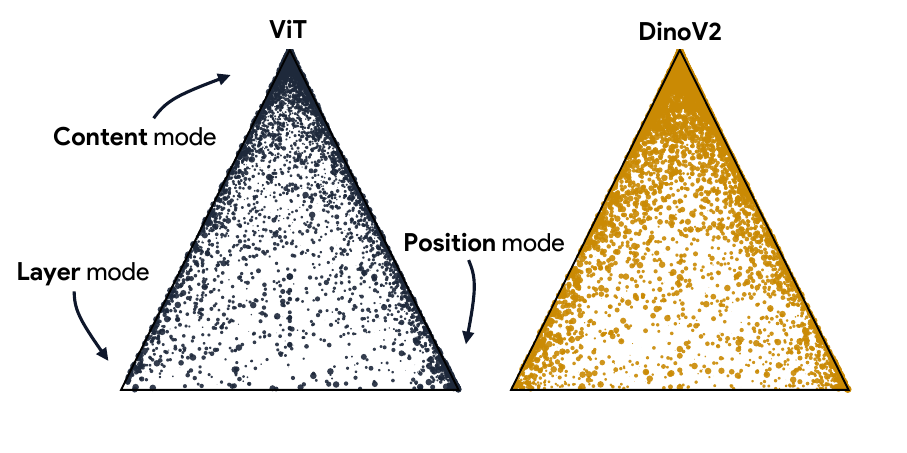}
    \caption{\textbf{Mode specialization across the network.} 
Each triangle (ViT on left and DINOv2 on right) visualizes all singular modes from all attention heads and all layers, plotted in barycentric coordinates indicating the relative contribution of layer, position, and content interaction families. Points near a vertex indicate modes dominated by layer-only, position-only, or content-only interactions while points along edges reflect mixed operator families (e.g., content–position). These layer-aggregated ternary plots summarize the overall computational footprint of different interactions in each model and highlight the characteristic content-centric organization induced by self-supervised training.}
    \label{fig:app:mode_specialization_points}
\end{figure}

\begin{figure*}[!t]
    \centering
    \includegraphics[width=0.99\linewidth]{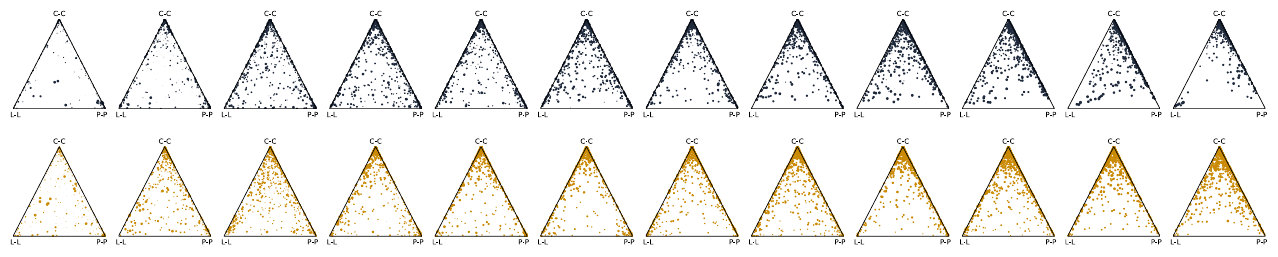}
    \caption{\textbf{Per-layer mode specialization.} 
Each triangle visualizes all singular modes from all attention heads in a given layer, plotted in barycentric coordinates indicating the relative contribution of layer, position, and content interaction families. Points near a vertex indicate modes dominated by layer-only, position-only, or content-only interactions while points along edges reflect mixed operator families (e.g., content–position). The top row shows ViT; the bottom row shows DINOv2. Across both models, modes exhibit strong functional specialization, with DINOv2 showing a broader distribution of content-dominated modes and more content–position operators in mid-layers.}
    \label{fig:app:mode_triangle_full}
\end{figure*}

\begin{figure*}
    \centering
    \includegraphics[width=0.99\linewidth]{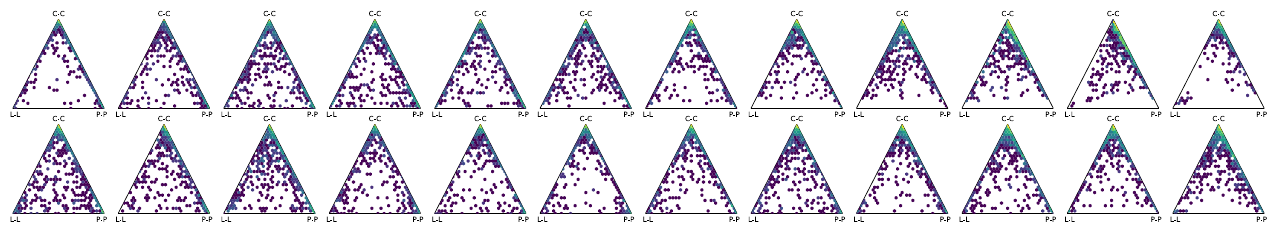}
    \caption{\textbf{Per-layer distribution of modes.} 
    Each triangle aggregates singular modes from all heads within a layer and displays their barycentric coordinates as a hexagonal density map. Lighter regions indicate higher density. ViT (top) and DINOv2 (bottom) show progressively increasing content bias with DINOv2 showing a richer mixture of content-position operators. These density plots reveal how the computational roles of modes evolve across depth at the level of the entire layer.}
    \label{fig:app:mode_triangle_full_density}
\end{figure*}

\paragraph{Layer aggregation and Density Plots.}
For each transformer layer $\ell$, we collect all singular modes across
its $H=12$ attention heads and $K=64$ modes per head, yielding a total of
$H\times K=768$ mode coordinates per layer:
\[
\mathcal{S}_\ell =
\big\{ (\hat{S}^{(i,h,\ell)}_{L},\,
        \hat{S}^{(i,h,\ell)}_{P},\,
        \hat{S}^{(i,h,\ell)}_{C})
        : i\in[1,K],\,h\in[1,H] \big\}.
\]

Each point corresponds to one singular mode’s relative energy
distribution among layer-, position-, and content-dominated
interactions. The per-layer ternary plots (Appendix \cref{fig:app:mode_triangle_full}) visualize these 768 points per
layer, showing how the informational specialization of modes evolves
with depth. To summarize the architecture as a whole, we pool all modes
across layers, heads, and factors ($12\times12\times64=9216$ points) and
plot their distribution in a single ternary simplex (\cref{fig:app:mode_specialization_points}), revealing each model’s overall specialization
bias.

To visualize specialization trends more clearly, we convert each
per-layer set of mode coordinates $\mathcal{S}_\ell$ into a continuous
density map within the ternary simplex. Each mode’s position
$(\hat{S}^{(i,h,\ell)}_{L},
  \hat{S}^{(i,h,\ell)}_{P},
  \hat{S}^{(i,h,\ell)}_{C})$
is treated as a sample from the layer’s specialization distribution.
We estimate a smooth kernel density over the simplex using hexagonal
binning (analogous to a 2D KDE) and normalize the resulting occupancy so
that densities sum to one within each layer. This produces the
color-coded plots in
\cref{fig:app:mode_triangle_full_density}, where \emph{lighter regions}
indicate a higher concentration of modes with similar informational
composition. These per-layer densities make it possible to track how
factor specialization sharpens or redistributes with depth—e.g., early
layers show broader mixtures, while deeper layers in DINOv2 cluster
tightly toward content- or position-dominated vertices.

\begin{figure*}
  \centering
  \includegraphics[width=\linewidth]{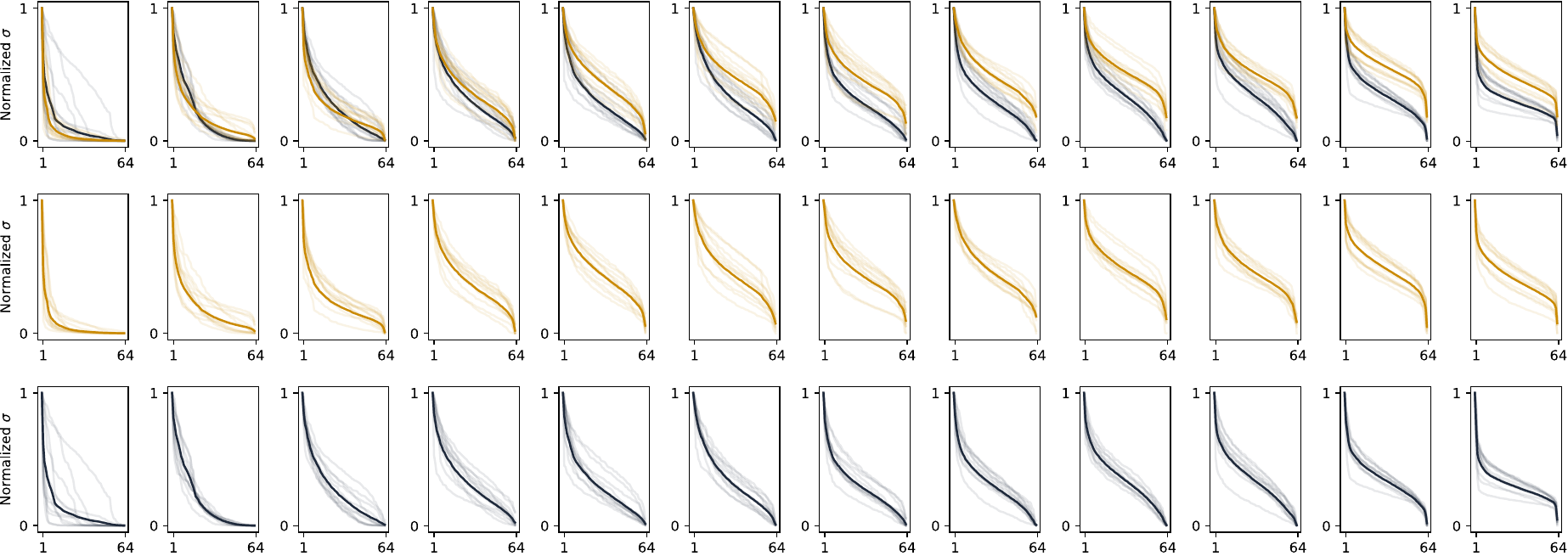}
  \caption{Details of $W$ singular values (normalized) across layers; ViT top, DINO bottom.}
  \label{fig:app:details_singular_value_w}
\end{figure*}

\paragraph{Spectra.}
We additionally visualize the singular value spectra for every head and
layer (Appendix \cref{fig:app:details_singular_value_w}). While the ViT spectra decay
rapidly, again indicating that only a few modes dominate each head, DINOv2
retains a flatter decay profile across the entire depth of the network.
This architectural difference implies that self-supervised training
supports more active communication channels throughout the model. Together, these visualizations expand the analysis of \hyperref[par:infoflow]{our characterization of information flow} by showing how informational factors are distributed
within each head and how this structure evolves across depth, providing a
complete head-and-mode resolution of information flow in both models.

\FloatBarrier
\clearpage

\clearpage
\section{Position representation}
\label{app:position_representation}

\begin{figure*}
    \centering
    \includegraphics[width=0.99\linewidth]{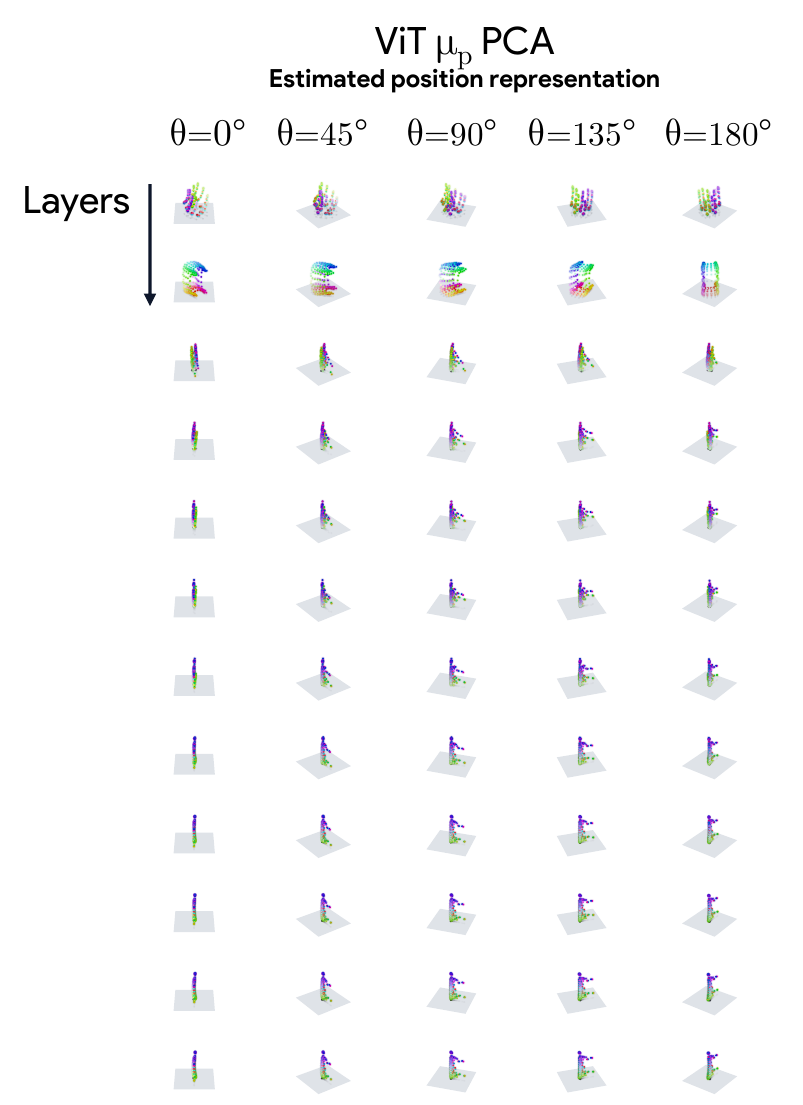}
    \caption{Estimated Position ($\mu_p$) in ViT visualized via 3D PCA under 5 rotation angles from layer 1 (top) to layer 12 (bottom).}
    \label{fig:app:details_position_ViT}
\end{figure*}

\begin{figure*}
    \centering
    \includegraphics[width=0.99\linewidth]{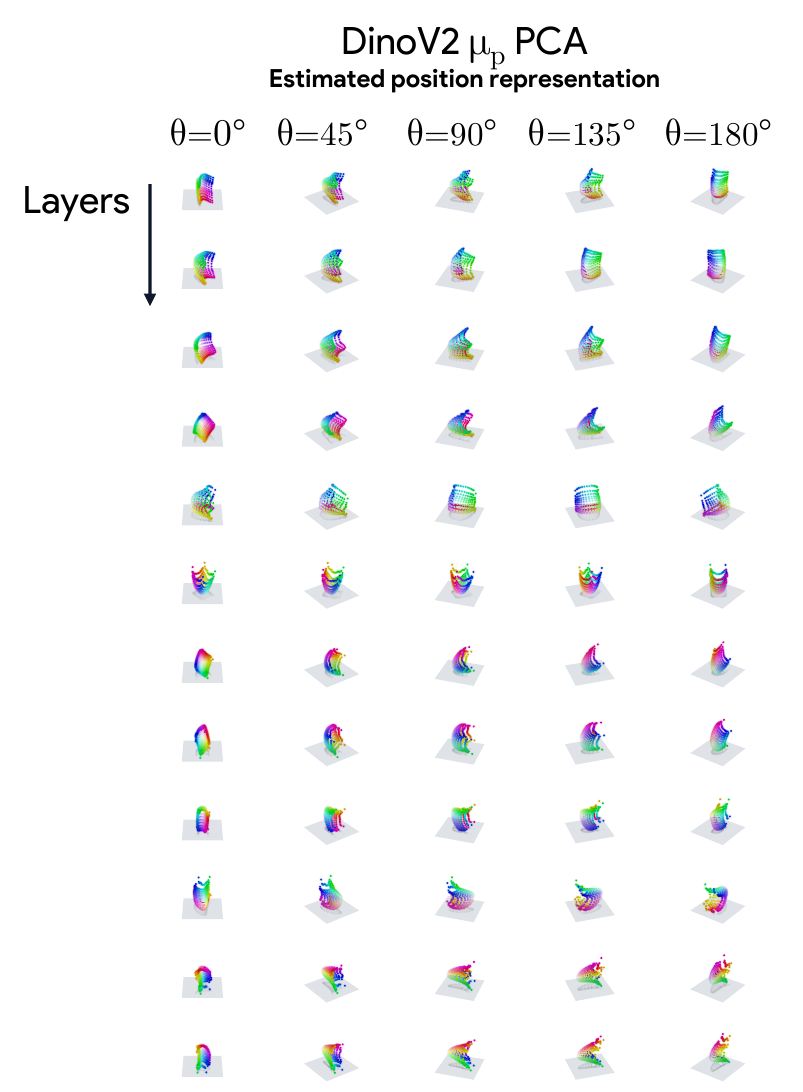}
    \caption{Estimated Position ($\mu_p$) in DINOv2 visualized via 3D PCA under 5 rotation angles from layer 1 (top) to layer 12 (bottom).}
    \label{fig:app:details_position_DINOv2}
\end{figure*}

~\cref{fig:app:details_position_ViT} and~\cref{fig:app:details_position_DINOv2} provide an expanded version of the positional PCA analysis introduced in
Figure~\ref{fig:pca_position}. For each layer $\ell$ and each model, we estimate the positional factor
$\position$ for all spatial tokens $\in \mathbb{R}^{196 \times 768}$ for ViT-B/16 (a $14 \times 14$ grid of tokens) and $\in \mathbb{R}^{256 \times 768}$ for DINOv2-B/14 (a $16 \times 16$ grid of tokens). We then run PCA directly on $\position$ and project each token onto
the top three principal components, yielding 3D coordinates in
$\mathbb{R}^{196 \times 3}$ and $\mathbb{R}^{256 \times 3}$. The resulting 3D point clouds are rendered from
five viewing angles
$\theta \in \{0^\circ, 45^\circ, 90^\circ, 135^\circ, 180^\circ\}$, so
that the geometry of the positional manifold can be inspected in 3D.

Each point is colored solely by its 2D spatial location using a fixed
position colormap (shown in the inset of ~\cref{fig:pca_position}): tokens are arranged on a
$14 \times 14$ or $16 \times 16$ grid and mapped to the corresponding
cell in the color chart (e.g., the bottom-right token, index $256$, uses
the dark blue color highlighted in the legend). This color coding is
identical across layers and models, allowing us to visually track how the
2D positional grid is embedded and distorted in the PCA space.

For the supervised ViT (~\cref{fig:app:details_position_ViT}), the positional
manifold quickly collapses with depth: by mid-layers the tokens cluster
along an effectively one-dimensional curve, and the 2D grid structure
becomes indistinguishable from most viewpoints. This matches the quasi-1D
structure already visible in the top panel of ~\cref{fig:pca_position}, and shows that
the collapse is robust to viewing angle rather than an artifact of a
particular projection. In contrast, DINOv2 (\cref{fig:app:details_position_DINOv2})
maintains an approximately planar 2D sheet across all layers: rotating the
view reveals that neighboring tokens remain embedded in a smooth, grid-like
surface, and no comparable 1D collapse occurs even at depth. These
complementary visualizations therefore substantiate our claim that
self-supervised training preserves an explicit 2D positional scaffold,
whereas supervised training compresses positional geometry to a much lower
effective dimensionality.

\FloatBarrier
\clearpage

\clearpage
\section{Enrichment of Content Across Layers}
\label{app:content_enrichment}

To quantify how token representations evolve across depth, we compute
pairwise layer--layer Pearson correlations between the flattened representations, both for the unfactorized block activations and for the content factor. This analysis characterizes how the geometry of patch representations changes across layers.

For each model, layer $\ell \in \{0,\dots,L{-}1\}$, and image $\x$, we extract either the unfactorized block activations $A_{\ell}(\x) \in \mathbb{R}^{T \times D}$ or the content factor $\mu_{c,\ell}(\x) \in \mathbb{R}^{T \times D}$, where $T$ is the number of total tokens (patch and the special tokens) and $D$ is the embedding dimension. We then flatten the representation into a vector $\mathbf{h}_{\ell}(\x) \in \mathbb{R}^{TD}$ and compute all pairwise Pearson correlations between these vectors,
yielding an $L \times L$ matrix. Averaging these matrices over all $N = 5{,}000$ images produces a
single layer--layer similarity matrix for each model. High correlations along the diagonal indicate representational stability across adjacent layers, whereas high off-diagonal correlations indicate that distant layers preserve similar token content. Lower off-diagonal values instead reflect progressive transformation of the representations with depth. Empirically, DINOv2 exhibits smoother mid-layer evolution than supervised ViT , consistent with the richer content operations revealed by our mode-level analysis.

\FloatBarrier
\clearpage

\end{document}